%% file: main.tex
\documentclass{article}

\input{math_commands.tex}

\usepackage{hyperref}
\usepackage{url}
\usepackage{graphicx}
\usepackage{caption}
\usepackage{xcolor}
\usepackage{tikz}
\usepackage{amsmath}
\usepackage{amssymb}
\usepackage{gensymb}
\usepackage{booktabs}  
\usepackage{tabularx}
\usepackage{float}

\usepackage{pifont}
\usepackage{microtype}
\usepackage{graphicx}
\usepackage{subfigure}
\usepackage{scalefnt}
\usepackage{enumitem}

\usepackage[accepted]{icml2021}

\definecolor{ADR_green}{RGB}{2,158,115}
\definecolor{Covar_orange}{RGB}{222,143,5}

\newcommand{\benchname}{\textit{TeachMyAgent}}
\begin{document}

\twocolumn[
\icmltitle{TeachMyAgent: a Benchmark for Automatic Curriculum Learning in Deep RL}



\icmlsetsymbol{equal}{*}

\begin{icmlauthorlist}
\icmlauthor{Cl\'ement Romac}{equal,inria}
\icmlauthor{R\'emy Portelas}{equal,inria}
\icmlauthor{Katja Hofmann}{microsoft}
\icmlauthor{Pierre-Yves Oudeyer}{inria}
\end{icmlauthorlist}

\icmlaffiliation{inria}{Inria, France}
\icmlaffiliation{microsoft}{Microsoft Research, UK}

\icmlcorrespondingauthor{Cl\'ement Romac}{clement.romac@inria.fr}
\icmlcorrespondingauthor{R\'emy Portelas}{remy.portelas@inria.fr}

\icmlkeywords{Curriculum Learning, Teacher Algorithms, Deep Reinforcement Learning, Benchmark}

\vskip 0.3in
]



\printAffiliationsAndNotice{\icmlEqualContribution} 

\begin{abstract}
Training autonomous agents able to generalize to multiple tasks is a key target of Deep Reinforcement Learning (DRL) research.
%
In parallel to improving DRL algorithms themselves, Automatic Curriculum Learning (ACL) study how teacher algorithms can train DRL agents more efficiently by adapting task selection to their evolving abilities.
%
While multiple standard benchmarks exist to compare DRL agents, there is currently no such thing for ACL algorithms. Thus, comparing existing approaches is difficult, as too many experimental parameters differ from paper to paper.
In this work, we identify several key challenges faced by ACL algorithms. Based on these, we present \benchname~(TA), a benchmark of current ACL algorithms leveraging procedural task generation. It includes 1) challenge-specific unit-tests using variants of a procedural Box2D bipedal walker environment, and 2) a new procedural Parkour environment combining most ACL challenges, making it ideal for global performance assessment.
We then use \benchname~to conduct a comparative study of representative existing approaches, showcasing the competitiveness of some ACL algorithms that do not use expert knowledge. We also show that the Parkour environment remains an open problem.
We open-source our environments, all studied ACL algorithms (collected from open-source code or re-implemented), and DRL students in a Python package available at {\small\url{https://github.com/flowersteam/TeachMyAgent}}.
\end{abstract}

\section{Introduction}
\label{sec:intro}

When looking at how structured and gradual human-learning is, one can argue that randomly presenting tasks to a learning agent is unlikely to be optimal for complex learning problems. Building upon this, curriculum learning has long been identified as a key component for many machine learning problems \citep{selfridge,elman,bengiocl,cangelosi2015developmental} in order to organize samples showed during learning.
While such a curriculum can be hand-designed by human experts on the problem, the field of Automatic Curriculum Learning \citep{graves2017automated,portelas2020-acl-drl} focuses on designing teacher algorithms able to autonomously sequence learning problem selection so as to maximize agent performance (e.g. over a set of samples in supervised learning, or game levels in DRL). 

Parallel to these lines of works, DRL researchers have been increasingly interested in finding methods to train generalist agents \citep{rajeswaran2016epopt,zhang2018study,surveymetarl,coinrun} to go beyond initial successes on solving single problems, e.g individual Atari games \citep{dqn} or navigation in fixed scenarios \citep{ddpg,sac}. Many works proposed novel DRL learning architectures able to successfully infer multi-purpose action policies when given an experience stream composed of randomly sampled \textit{tasks} \citep{uvfa,rainbow,coinrun,Hessel-2019-popart}. Here and thereafter \textit{tasks} denote learning problems in general, for instance multiple mazes to solve (a.k.a environments) or, in the context of robotic manipulation, multiple state configuration to obtain (a.k.a. goals) \citep{portelas2020-acl-drl}. To compare existing and future Multi-task DRL agents, \citet{procgen} proposed a suite of 16 atari-like environments, all relying on Procedural Content Generation (PCG) to generate a wide diversity of learning situations. The high-diversity induced by PCG has been identified as particularly beneficial to foster generalization abilities to DRL agents \citep{illuminating,risiPCG,OpenAI2019SolvingRC}.


 An important aspect not covered by these prior works is that they all rely on proposing randomly selected tasks to their agent, i.e. they do not consider using curriculum in learning. One can argue that random task selection is inefficient, especially when considering complex continuous task sets, a.k.a task spaces, which can feature subspaces of varying difficulties ranging from trivial to unfeasible. Following this observation, many works attempted to train given multi-task agents by pairing them with ACL algorithms \citep{portelas2020-acl-drl}. The advantages of ACL over random task sampling for DRL agents have been demonstrated in diverse experimental setups, such as domain randomization for sim2real robotics \citep{OpenAI2019SolvingRC,ADRmila}, video games \citep{montezuma-single-demo,tscllike}, or navigation in procedurally generated environments \citep{goalgan,portelas2019,settersolver}.

While this diversity of potential application domains and implementations of ACL hints a promising future for this field, it also makes comparative analysis complicated, which limits large-scale adoption of ACL. For instance, depending on the ACL approach, the amount of required expert knowledge on the task space can range from close to none -- as in \citet{portelas2019} -- to a high amount of prior knowledge, e.g. initial task sampling subspace and predefined reward range triggering task sampling distribution shifts, as in \citet{OpenAI2019SolvingRC}. Additionally, some ACL approaches were tested based on their ability to master an expert-chosen target subspace \citep{selfpaceddrl} while others were tasked to optimize their performance over the entire task space \citep{riac,goalgan,portelas2019}. Besides, because of the large computational cost and implementation efforts necessary for exhaustive comparisons, newly proposed ACL algorithms are often compared to only a subset of previous ACL approaches \cite{ADRmila,portelas2019,settersolver}. This computation bottleneck is also what prevents most works from testing their ACL teachers on a diversity of DRL students, i.e. given a set of tasks, they do not vary the student's learning mechanism nor its embodiment.
Designing a unified benchmark platform, where baselines would be shared and allow one to only run its approach and compare it to established results, could drive progress in this space.

\begin{figure}[ht]
\vskip 0.2in
\begin{center}
\centerline{\includegraphics[width=\columnwidth]{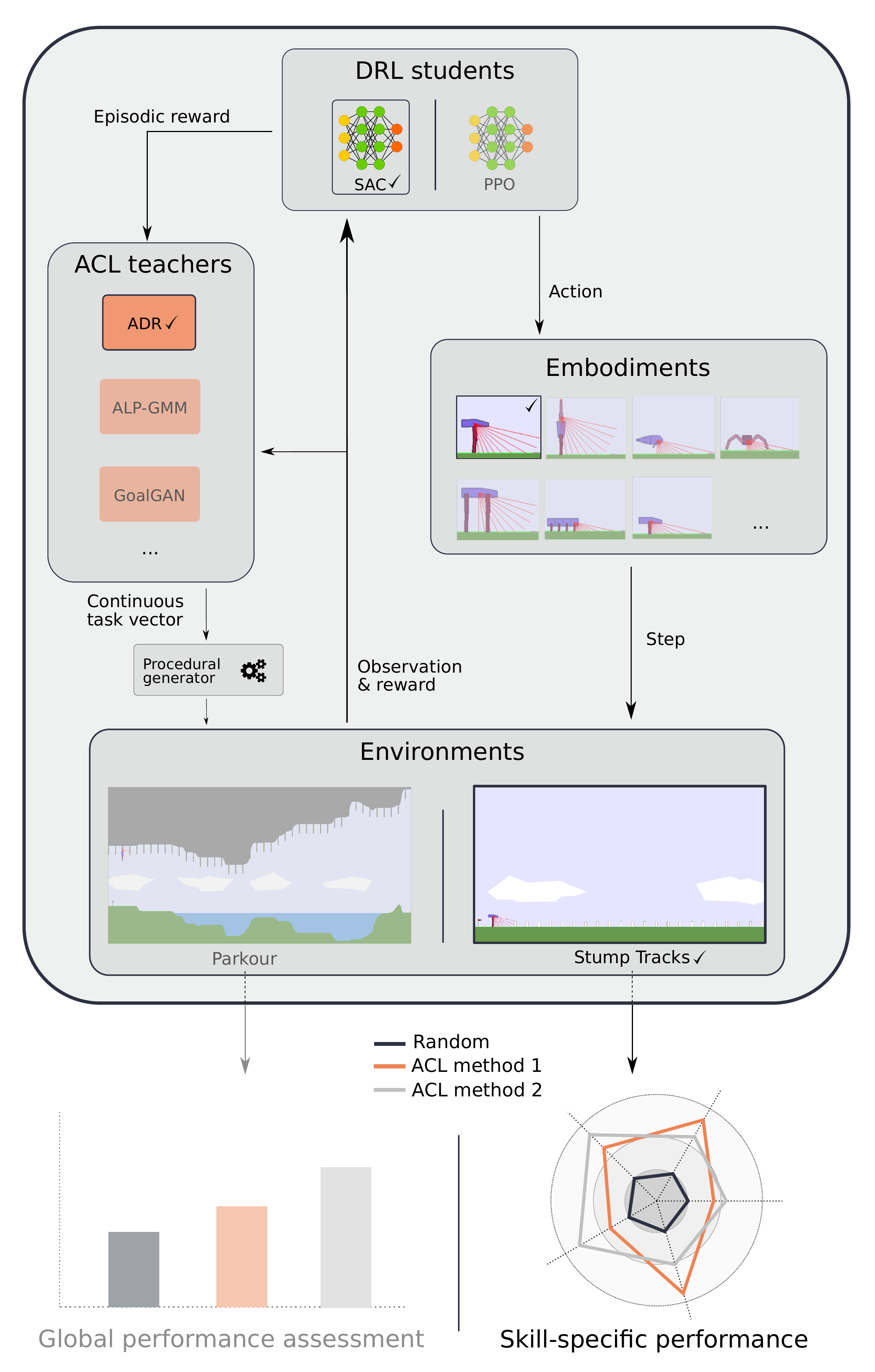}}
\caption{\textbf{TeachMyAgent}: A benchmark to study and compare teacher algorithms in continuous procedural environments.}
\label{fig:bench-vizu}
\end{center}
\vskip -0.2in
\end{figure}

Inspired by how the MNIST dataset \citep{mnist} or the ALE Atari games suite \citep{ALE-bench} respectively catalyzed supervised learning and single-task reinforcement learning research, we propose to perform this much-needed in-depth ACL benchmarking study. As such, we introduce \benchname~\textit{1.0}\footnote{\footnotesize\scalefont{0.96}\url{http://developmentalsystems.org/TeachMyAgent/}}, a teacher testbed featuring a) two procedural Box2D\footnote{2D game engine, used in OpenAI gym \citep{gym}} environments with challenging task spaces, b) a collection of pre-defined agent embodiments, and c) multiple DRL student models. The combination of these three components constitutes a large panel of diverse teaching problems. We leverage this benchmark to characterize the efficiency of an ACL algorithm on the following key teaching challenges:
\begin{enumerate}[itemsep=6pt,parsep=0pt,topsep=0pt]
    \item \textit{Mostly unfeasible task spaces} - While using PCG systems to generate tasks allows to propose rich task spaces to DRL agents, which is good for generalization, such large spaces might contain a predominant amount of unfeasible (or initially unfeasible) tasks . A teacher algorithm must then have the ability to quickly detect and exploit promising task subspaces for its learner.
     \item \textit{Mostly trivial task spaces} - On the contrary, the task space might be mostly trivial and contain only few challenging subspaces, which is a typical scenario when dealing with a skilled student (e.g. that is already trained, or that has an advantageous embodiment). In that case the teacher has to efficiently detect and exploit the small portion of subspaces of relevant difficulty.
     \item \textit{Forgetting students} - DRL learners are prone to catastrophic forgetting \cite{Kirkpatrickforgetting}, i.e. to overwrite important skills while training new ones. This has to be detected and dealt with by the teacher for optimal curriculum generation. 
     \item \textit{Robustness to diverse students} - Being able to adapt curriculum generation to diverse students is an important desiderata to ensure a given ACL mechanism has good chances to transfer to novel scenarios.
     \item \textit{Rugged difficulty landscapes} - Another important property for ACL algorithms is to be able to deal with task spaces for which the optimal curriculum is not a smooth task distribution sampling drift across the space but rather a series of distribution jumps, e.g. as in complex PCG-task spaces.
     \item \textit{Working with no or little expert knowledge} - Prior knowledge over a task space w.r.t. a given student is a costly information gathering process that needs to be repeated for each new problem/student. Relying on as little expert knowledge as possible is therefore a desirable property for ACL algorithms (especially if aiming for out-of-the-lab applications).
\end{enumerate}
To precisely assess the proficiency of an ACL algorithm on each of these challenges independently, we extend a Box2D walker environment from \citet{portelas2019} into multiple unit-test variants, one per challenge, inspired by the structure of \textit{bsuite}  \citep{bsuite2019}, a recent benchmark for RL agents. The second environment of our benchmark is the \textit{Parkour} environment, inspired by \citet{poet2}. It features a complex task space whose parameters seed a neural network-based procedural generation of a wide diversity of environments, in which there exists drastically different learning curricula depending on the agent's embodiment (see fig. \ref{fig:bench-vizu}). To assess the ability of existing ACL methods to robustly adapt to diverse students, we consider a random black-box student scenario in the Parkour environment, i.e. the morphology (e.g. walker or climber) of the learner is randomly selected for each new training run.

 \paragraph{Scope} More precisely, we conduct an in-depth comparative study of ACL approaches suited for generalist DRL agents such as SAC \citep{sac} or PPO \citep{ppo} in single agent scenarios. We do not include works on self-play/multi-agent setups \cite{self-play-framework,ma-survey} nor single-agent population-based approaches \citep{imgep,poet2}. Also, we are interested in the problem of task selection from a continuous parameter space encoding the procedural generation of tasks. We leave the analysis of ACL methods for discrete task sets \cite{tscl,tscllike}, sets of task spaces \cite{imgep,curious}, or intrinsic reward learning \citep{icm,rnd} for future work. We assume this continuous space is given and relatively low-dimensional as it already poses strong teaching challenges: we therefore leave the analysis of approaches that autonomously learn task representations for subsequent work \cite{skewfit,metarl-carml,grimgep}.

Our main contributions are:
\begin{itemize}[itemsep=6pt,parsep=0pt,topsep=0pt]
    \item Identification of multiple challenges to be tackled by ACL methods, enabling multi-dimensional comparisons of these algorithms.
    \item \benchname~\textit{1.0}, a set of teaching problems (based on PCG environments) to study and compare ACL algorithms when paired with DRL students.
    \item Comparative study of representative existing ACL approaches including both skill-specific unit-tests and global performance assessments, which highlights the competitiveness of methods not using expert knowledge and shows that our Parkour environment largely remains an open problem for current state-of-the-art ACL.
    \item Release of an open-source Python package, featuring 1) all environments, embodiments and DRL students from \benchname, 2) all studied ACL algorithms, that we either adapt to our API when code is available or re-implement from scratch if not open-sourced, 3) our experimental results as baselines for future works, and 4) tutorials \& reproducibility scripts.
\end{itemize}


\section{Related work}

Many environment suites already exist to benchmark DRL algorithms: some of them leverage video games, which provide challenging discrete action spaces, e.g. Atari 2600 Games as in \citet{ALE-bench} or Sonic The Hedgehog levels in \citet{sonic-bench}. To study and develop DRL agents suited for complex continuous control scenarios, the community predominantly used the MuJoCo physics engine \citep{mujoco}. The Deep Mind Lab \citep{DMlab} provides customizable puzzle-solving environment, particularly well suited to study goal-conditioned policies learning from pixels in rich 3D environments. At the intersection of DRL and Natural Language Processing, benchmark environments such as TextWorld \citep{textworld} or BabyAI \citep{BAbyAI} were also designed to provide a testbed to develop autonomous agent receiving linguistic goals and/or interacting using language. The \textit{bsuite} benchmark \citep{bsuite2019} leverages unit-tests to assess the core capabilities of DRL methods (e.g. generalization, memory). In all these previous works, the DRL agent is learning in one or few environments presented randomly and/or intrinsically chooses goals within those predefined environments, and the long-term community objective is to find more efficient learning architectures. On the contrary, the objective of \benchname~ is to foster the development of new teacher algorithms whose objective is, given a task space and a DRL student, to most efficiently organize the learning curriculum of their DRL student such that its performance is maximized over the task set. In other words, it is not about finding efficient learning architectures but about finding efficient curriculum generators.

Perhaps closest to our work is the Procgen benchmark \citep{procgen}, which features several atari-like environments, all having unique procedural generation systems allowing to generate a wide diversity of learning situations, particularly well suited to assess the generalization abilities of DRL agents. While they rely on an uncontrolable, random procedural generation, we assume control over it, which enables the use of ACL methods to select parameters encoding task generation. An interesting future work, parallel to ours, would be to modify the Procgen benchmark to allow direct control over the procedural generation.

Because of the current lack of any ACL benchmark, most recently proposed ACL algorithms relied on designing their own set of test environments. \citet{goalgan} used a custom MuJoCo Ant maze in which the ACL approach is in control of which end-position to target. \citet{selfpaceddrl} used another MuJoCo Ant maze and ball-catching environment featuring a simulated Barrett WAM robot. While these previous works studied how to control goal selection in a given fixed environment, we are interested in the arguably more challenging problem of controlling a rich parametric procedural generation. \citet{portelas2019} already studied ACL in Stump Tracks, a procedural Box2D environment that we include and extend in \benchname, however it did not perform an extensive comparative study as what we propose in the present work. \citet{settersolver} also used procedural generation to test their ACL approach, however they only compared their ACL algorithm to Goal-GAN \citep{goalgan}, and did not open-source their environments. Additionally, in contrast with all previously cited ACL works, in \benchname~ we propose an in-depth analysis of each approaches through multiple unit-test experiments to fully characterize each teacher.

\section{ACL baselines}
In the following paragraphs we succinctly frame and present all the ACL algorithms that we compare using \benchname. More detailed explanations are left to appendix \ref{app:acl-details}.

\paragraph{Framework} Given a DRL student $s$ and a n-dimensional task-encoding parameter space $\mathcal{T} \in \mathbb{R}^n$  (i.e. a task space), the process of Automatic Curriculum Learning aims to learn a function $\mathcal{A}: \mathcal{H} \mapsto \mathcal{D}(\mathcal{T})$ mapping any information retained about past interactions with the task space to a distribution of tasks.

One can define the optimization objective of an ACL policy given an experimental budget of $E$ episodic tasks as:
\begin{equation}
    \label{eq:acl-obj}
    \max_{\mathcal{A}} \int_{\mathrm{T}\sim \mathcal{D}_{target}} \! P_{\mathrm{T}}^E\, \mathrm{d}\mathrm{T},
\end{equation}
with $\mathcal{D}_{target}$ the distribution of test tasks over the task space and $P$ the post-training performance (e.g. episodic reward, exploration score) of student $s$ on task $\mathrm{T}$ after $E$ episodes. Since it is usually difficult to directly optimize for this objective, various surrogate objectives have been proposed in the literature. See \citet{portelas2020-acl-drl} for a review and classification of recent ACL works.

\paragraph{Expert-knowledge} To ease the curriculum generation process, multiple forms of expert knowledge have been provided in current ACL approaches. We propose to gather them in three categories: 1)  use of initial task distribution $\mathcal{D}_{init}$ to bootstrap the ACL process, 2) use of a target task distribution $\mathcal{D}_{target}$ to guide learning, and 3) use of a function interpreting the scalar episodic reward sent by the environment to identify mastered tasks (\textit{Reward mastery range}). For each implemented ACL method, we highlight its required prior knowledge over the task space w.r.t a given DRL agent in table \ref{acl-algo-table}. We hope that this classification will ease the process of selecting an ACL method for researchers and engineers, as available expert knowledge is (arguably) often what conditions algorithmic choices in machine learning scenarios.

\paragraph{Implemented baselines}
\label{sec:acl-algos}
We compare seven ACL methods, chosen to be representative of the diversity of existing approaches, that can be separated in three broad categories. First, we include three methods relying on the idea of maximizing the Learning Progress (LP) of the student: RIAC \citep{riac}, Covar-GMM \citep{moulinfriergmm} and ALP-GMM \citep{portelas2019}. We then add in our benchmark Goal-GAN \citep{goalgan} and Setter-Solver \citep{settersolver}, both generating tasks using deep neural networks and requiring a binary reward for mastered/not mastered tasks, pre-defined using expert knowledge. Finally, we append to our comparison two ACL algorithms using the idea of starting from an initial distribution of tasks and progressively shifting it regarding the student's capabilities: ADR \citep{OpenAI2019SolvingRC} (inflating a task distribution from a single initial task based on student mastery at each task distribution's border) and SPDL \citep{selfpaceddrl} (shifting its initial distribution towards a target distribution). We also add a baseline teacher selecting tasks uniformly random over the task space (called Random).


\begin{table}[t]
\caption{Expert knowledge used by the different ACL methods. We separate knowledge required (REQ.) by algorithms, optional ones (OPT.), and knowledge not needed (empty cell).}
\label{acl-algo-table}
\vskip 0.15in
\setlength\tabcolsep{1.4pt}
\begin{center}
\begin{small}
\begin{sc}
\begin{tabular}{l|cccc}
\toprule
\textbf{Algorithm} & $\mathcal{D}_{init}$ & $\mathcal{D}_{target}$ & Reward mastery range \\ \hline\hline
     ADR           & req. &            & req.  \\ \hline
     ALP-GMM       &  opt.   &            &             \\  \hline
     Covar-GMM     &  opt.   &            &             \\  \hline
     Goal-GAN      & opt. &          & req.  \\  \hline
     RIAC          &            &            &             \\  \hline
     SPDL    & req. & req. &                   \\  \hline
     Setter-Solver &            & opt. & req. \\ \bottomrule 
\bottomrule
\end{tabular}
\end{sc}
\end{small}
\end{center}
\vskip -0.1in
\end{table}

\section{The \benchname~benchmark}
In the following section, we describe available environments and learners in \benchname. We propose two Box2D environments with procedural generation allowing to generate a wide variety of terrains. Both our environments are episodic, use continuous action/observation spaces and return scalar rewards. In addition, we provide two DRL algorithms as well as multiple agent morphologies. An experiment is thus constituted of an ACL method, an environment and a learner (i.e. an embodied DRL algorithm).

\subsection{Environments}

\begin{figure}[ht]
\begin{center}
\centerline{\includegraphics[width=0.65\columnwidth]{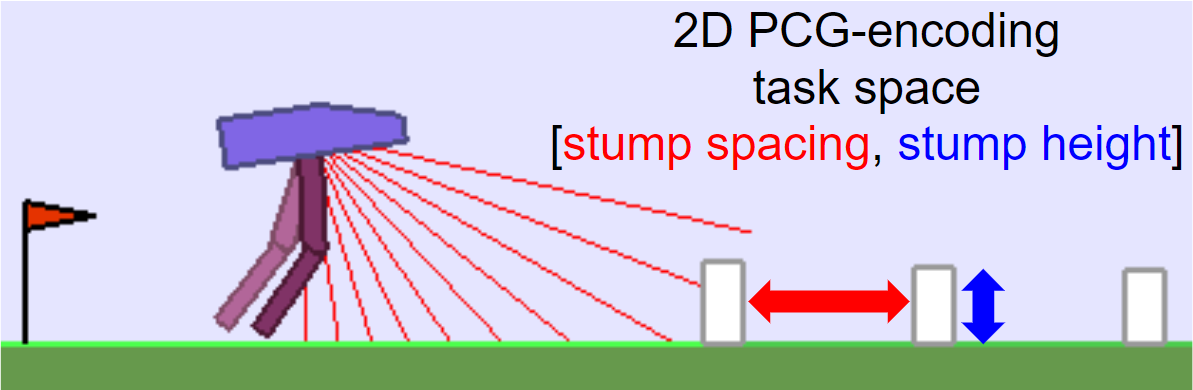}}
\caption{\textit{Stump Tracks}, a simple parametric env. to study ACL algorithms with DRL students.}
\label{fig:stump-tracks}
\end{center}
\vskip -0.3in
\end{figure}

\paragraph{Stump Tracks environment} Stump Tracks is an extension of a parametric Box2D environment initially presented in \citet{portelas2019}. The learning policy is embodied into a walker agent whose motors are controllable with torque (i.e. continuous action space). The observation space is composed of lidar sensors, head position and joint positions. The walker is rewarded for going forward and penalized for torque usage. An episode lasts $2000$ steps at most, and is terminated if the agent reaches the end of the track or if its head collides with the environment (in which case a $-100$ reward is received). A $2$D parametric PCG is used for each new episode: it controls the height and spacing of stumps laid out along the track (see fig. \ref{fig:stump-tracks} and app. \ref{app:env-details}). We chose to feature this environment as its low-dimensional task space is convenient for visualizations and modifications. We derive multiple variants of Stump Tracks (e.g. by extending the task space boundaries or shuffling it) to design our unit-tests of ACL challenges (see sec. \ref{sec:intro} and sec. \ref{sec:experiments}).

\paragraph{Parkour environment} Inspired by both Stump Tracks and another Box2D environment from \citet{poet2}, we present the parametric Parkour environment: a challenging task space with rugged difficulty landscape, few prior knowledge definable, and requiring drastically different learning curricula depending on the agent's embodiment.

It features an uneven terrain (see figure \ref{fig:bench-vizu}) composed of a ground and ceiling encoded through a Compositional Pattern-Producing Network (CPPN) \citep{CPPN}. This CPPN, whose weights and architecture are kept fixed, takes an additional input vector of bounded real numbers which acts as the parameters controlling terrain generation. This neural network based generation enables to create a task space with a rugged difficulty landscape (see appendix \ref{app:env-details}), requiring time consuming exploration from an expert to seek trivial subspaces. We propose three versions of this task space (i.e. three possible bounds for the CPPN's input vector): easy, medium (used in the experiments of this paper) and hard. 
The Parkour environment also features graspable objects, called "creepers", creating a niche for climbing morphologies. Similarly to the stumps in Stump Tracks, the creepers' generation is controlled by their height and the space between them.
The Parkour's task space also contains a dimension controlling the "water" level of the track, ranging from $0$ (no water) to $1$ (entire parkour under water). Water adds new physic rules aiming to imitate (in a simplified way) physics of water.

The resulting $6$D task space (3 for the CPPN's input, 2 for creepers and 1 for water) creates a rich environment in which the optimal curriculum will largely depend on the agent's embodiment (e.g. swimming agents need high levels of water, while climbers and walkers need low levels). Note that, as in Stump Tracks, each episode lasts $2000$ steps, agents are rewarded for moving forward (and penalised for using torque) and have access to lidars, head position, joint positions, and also additional information (see appendix \ref{app:env-details}).

\subsection{Learners}
\paragraph{Embodiments} As aforementioned, we introduce new morphologies using swimming and climbing locomotion (e.g. fish, chimpanzee, see figure \ref{fig:bench-vizu}). \benchname~ also features the short walker and quadrupedal walker from \citet{portelas2019} as well as new walking morphologies such as the spider and the millipede (see figure \ref{fig:bench-vizu}).

\vspace{-0.4cm}\paragraph{DRL algorithms} To benchmark ACL algorithms, we rely on two different state-of-the-art DRL algorithms: 1) Soft-Actor-Critic \citep{sac} (SAC), a now classical off-policy actor-critic algorithm based on the dual optimization of reward and action entropy, and 2) Proximal Policy Optimization (PPO) \citep{ppo}, a well-known on-policy DRL algorithm based on approximate trust-region gradient updates. We use OpenAI Spinningup's implementation\footnote{\url{https://spinningup.openai.com}} for SAC and OpenAI Baselines' implementation\footnote{\url{https://github.com/openai/baselines}} for PPO. See appendix \ref{app:expe-details} for implementation details.

\section{Experiments}\label{sec:experiments}

We now leverage \benchname~ to conduct an in-depth comparative study of the ACL algorithms presented in section \ref{sec:acl-algos}. After discussing experimental details, we undergo two separate experiments, aiming to answer the following questions: 
\begin{itemize}
    \item How do current ACL methods compare on each teaching challenges proposed in sec. \ref{sec:intro} ?
    \item How do current ACL methods scale to a complex task space with limited expert knowledge ?
\end{itemize}

\subsection{Experimental details}
For both our environments, we train our DRL students for $20$ million steps. For each new episode, the teacher samples a new parameter vector used for the procedural generation of the environment. The teacher then receives the cumulative episodic reward that can be potentially turned into a binary reward signal using expert knowledge (as in GoalGAN and Setter-Solver). Additionally, SPDL receives the initial state of the episode as well as the reward obtained at each step, as it is designed for non-episodic RL setup. Every $500000$ steps, we test our student on a test set composed of $100$ pre-defined tasks and monitor the percentage of test tasks on which the agent obtained an episodic reward greater than $230$ (i.e. "mastered" tasks), which corresponds to agents that were able to reach the last portion of the map (in both Stump Tracks and Parkour). We compare performance results using Welch's t-test as proposed in \citet{colas_seeds_2018}, allowing us to track statistically significant differences between two methods. We perform a hyperparameter search for all ACL conditions through grid-search (see appendix \ref{app:acl-details}), while controlling that an equivalent number of configurations are tested for each algorithm. See appendix \ref{app:expe-details} for additional experimental details.

\subsection{Challenge-specific comparison with Stump Tracks} 
First, we aim to compare the different ACL methods on each of the six challenges we identified and listed in section \ref{sec:intro}. For this, we propose to leverage the Stump Tracks environment to create five experiments, each of them designed to highlight the ability of a teacher in one the first five ACL challenges (see appendix \ref{app:expe-details} for details): 
\begin{itemize}
    \item \textit{Mostly unfeasible task space}: growing the possible maximum height of stumps, leading to almost $80\%$ of unfeasible tasks.
    \item \textit{Mostly trivial task space}: allowing to sample stumps with negative height introducing $50\%$ of new trivial tasks.
    \item \textit{Forgetting student}: resetting the DRL model twice throughout learning (i.e. every $7$ Millions steps).
    \item \textit{Diverse students}: using multiple embodiments (short bipedal and spider) and DRL students (SAC and PPO).
    \item \textit{Rugged difficulty landscape}: Applying a random transformation to the task space such that feasible tasks are scattered across the space (i.e. among unfeasible ones).
\end{itemize}


Additionally, in order to compare methods on the last challenge (i.e. the need of prior knowledge), we propose to perform each of our five experiments in three conditions: 
\begin{itemize}
    \item \textit{No expert knowledge}: None of the prior knowledge listed in table \ref{acl-algo-table} is given. Hence only methods not requiring it can run in this setup.
    \item \textit{Low expert knowledge}: Only reward mastery range information is accessible. We consider this as low prior knowledge as, while it requires some global knowledge about the task space, it does not require assumptions on the difficulty of specific subspaces of the task space.
    \item \textit{High expert knowledge}: All the expert knowledge listed in table \ref{acl-algo-table} is given. 
\end{itemize}

Note that in the \textit{No expert knowledge} and \textit{Low expert knowledge} setups, SPDL (and ADR in \textit{Low expert knowledge}) uses an initial task distribution randomly chosen as a subset of the task space. Moreover, in order to make a fair comparison in the\textit{ High expert knowledge} condition, we modified the vanilla version of Covar-GMM and ALP-GMM such that they can use an expert-given initial task distribution.

Using these $15$ experiments ($5$ challenges in $3$ expert knowledge setups), we here introduce what is, to our knowledge, the first unit-test like experiment of ACL methods, allowing one to compare teachers in each of the challenges we previously introduced. Moreover, performing each of the five experiments in three expert knowledge setups allows to show how the (un)availability of expert knowledge impacts performance for each method, which is hard to infer from each approach's original paper as they tend to focus only on the most ideal scenario. See appendix \ref{app:expe-details} for a detailed explanation of each experimental setup. 

\begin{figure*}[htb]
\vskip 0.2in
\begin{center}
\centerline{\includegraphics[width=\textwidth]{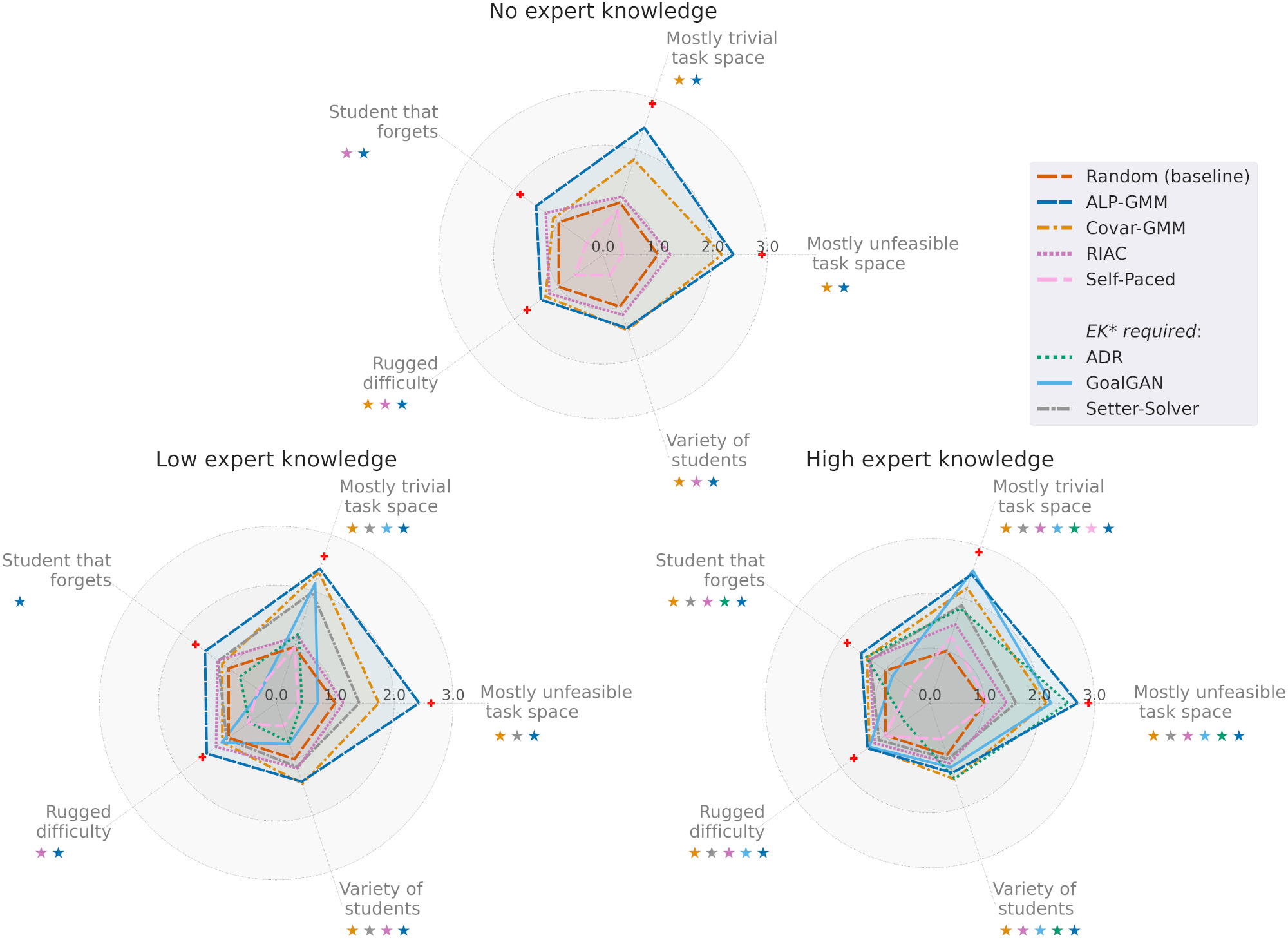}}
\caption{\textit{EK: Expert Knowledge}. Post-training performance of each ACL method as a ratio of Random's results on multiple teaching challenges, done with $3$ different expert knowledge levels. We use {\textcolor{red}{\ding{58}}} to show estimations of upper-bound performances in each challenge, except for \textit{Variety of students} (see appendix \ref{app:additional-results_original-stump-tracks}). On each axis, we indicate which method performed significantly better than Random ($p<0.05$) using colored stars matching each method's color (e.g. {\textcolor{Covar_orange}{\ding{72}}} for Covar-GMM, {\textcolor{ADR_green}{\ding{72}}} for ADR). See appendix \ref{app:additional-results_stump-tracks} for details.}
\label{fig:radar_chart}
\end{center}
\vskip -0.2in
\end{figure*}

To conduct our analysis, each ACL method is used in $15$ experiments with $32$ seeds, except ADR, GoalGAN and Setter-Solver which cannot run in the \textit{No expert knowledge} setup (i.e. only $10$ experiments). We then calculate the aforementioned percentage of mastered test tasks on our test set (identical for all experiments), and average it over seeds. Performance results of all conditions can be visualized in figure \ref{fig:radar_chart} as a ratio of the Random teachers' performance, our lower-baseline.


 \paragraph{Results} We gather the results in figure \ref{fig:radar_chart} as well as in appendix \ref{app:additional-results_stump-tracks}. 
 
 \textit{Expert-knowledge-free methods --~~} Using these, one can see, first, that methods not requiring any expert knowledge (e.g. ALP-GMM or Covar-GMM) obtain very similar performances in \textit{No expert knowledge} and in \textit{High expert knowledge} setups (although expert knowledge does benefit them in terms of sample efficiency, see  app. \ref{app:additional-results_stump-tracks} for details). Comparing their performance without prior knowledge to the results obtained by other teachers when they have access to high expert knowledge shows how competitive expert-knowledge-free methods can be.

 \textit{Expert knowledge dependency --~~} The \textit{Low expert knowledge} setup highlights the dependence of methods relying on an initial distribution of easy tasks (e.g. ADR and GoalGAN), as it is not given in this scenario. As a result, in this setup, ADR obtains end performances not significantly different from Random in all challenges, and GoalGAN only outperforms Random in the mostly trivial task space ($p<0.05$). This has to be compared with their performance on the \textit{High expert knowledge} setup, in which both approaches reach the top 3 results on 3/5 challenges.

\textit{ADR \& GoalGAN --~~} Both ADR and GoalGAN have one strong weakness in a challenge (\textit{Rugged difficulty} for ADR and \textit{Forgetting student} for GoalGAN) that lead them to a performance worse than Random (significantly for ADR with $p<0.05$) in all expert knowledge setups. For ADR, it can be explained by the fact that its expansion can get stuck by subspaces of very hard (or unfeasible) difficulty, and for GoalGAN, by its inability to adapt quickly enough to the student's regressing capabilities because of its inertia to update its sampling distribution (updating the buffer and training the GAN). We provide a more in-depth analysis of these two cases in appendix \ref{app:additional-results_stump-tracks}.

 \textit{SPDL --~~} One can see that SPDL's performance seem very poor in our experimental setup: its end performance is significantly inferior to Random in 11/15 experiments ($p<0.05$). 
 This can be explained by the fact that SPDL, by design, optimizes performance over a Gaussian target distribution, while our test set is uniformly sampled over the task space.
 See appendix \ref{app:acl-details} for details and potential fixes.
 

\begin{figure*}[htb]
\vskip 0.2in
\begin{center}
\centerline{\includegraphics[width=\textwidth]{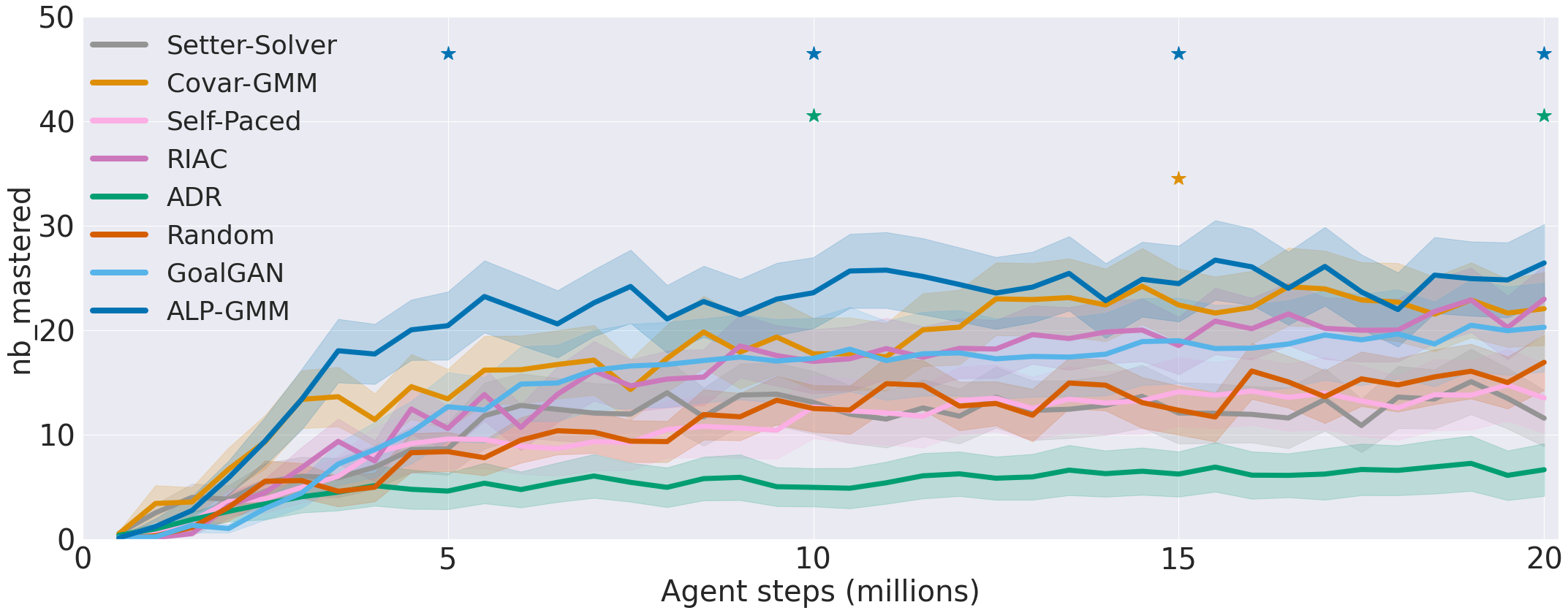}}
\caption{Averaged performance ($48$ seeds, with standard error of the mean) for each ACL method on Parkour. We calculate every $5$ millions steps which method obtained statistically different ($p<0.05$) results from Random and indicate it with a star.}
\label{fig:parkour_comparison}
\end{center}
\vskip -0.2in
\end{figure*}
 
\subsection{Global performance analysis using the Parkour}
The second experiment we propose aims to more broadly benchmark ACL methods' performance in the Parkour environment, which features most of the previously discussed ACL challenges: 1) most tasks are unfeasible, 2) before each run, unknown to the teacher, the student's embodiment is uniformly sampled among three morphologies (bipedal walker, fish and chimpanzee), requiring the teacher to adapt curriculum generation to a diversity of student profiles, and 3) tasks are generated through a CCPN-based PCG, creating a rich task space with rugged difficulty landscape and hardly-definable prior knowledge (see appendix \ref{app:env-details}).

We perform $48$ seeded experiments (i.e.\ $16$ seeds per morphology). To evaluate performance, three test sets were hand-designed (one per embodiment) such that each contains an even distribution between easy, medium and hard tasks. In terms of expert knowledge for teachers, we only give reward mastery range. Without straightforward initial easy task distribution to give to teachers requiring such knowledge (ADR and SPDL), we set it randomly over the space for each new run. See appendix \ref{app:expe-details} for details. 



 
\paragraph{Results} We present the evolution of performance of each teacher averaged over all seeds (and thus all embodiments) in figure \ref{fig:parkour_comparison} and gather the detailed results in appendix \ref{app:additional-results_parkour}.
Interestingly, one can observe that best-performing methods do not use expert knowledge. This is explained by the fact that few prior knowledge is provided to the teachers in these experiments and, as shown in the challenge-specific experiments, most methods using expert knowledge heavily rely on them to reach high-performance.
However, one can see that, while SPDL and Setter-Solver remain at the performance level of Random, GoalGAN's performance along training is (mostly) not significantly different from those of Covar-GMM and RIAC, two methods not relying on expert knowledge, as opposed to GoalGAN.
On his side, ADR seems to plateau very fast and finally reach an average performance significantly worse than Random ($p<0.05$, see figure \ref{fig:full_parkour_comparison}). Indeed, as the difficulty landscape of the Parkour environment is rugged, and the initial "easy" task distribution randomly set, ADR is unable to progressively grow its sampling distribution towards feasible subspaces. 
Finally, when looking specifically to each embodiment type, results show the incapacity of all teachers to make the DRL student learn an efficient policy with the climbing morphology (i.e. at most $1\%$ of mastered tasks by the end of training across all teachers), although we are able to show that high-performing policies can be learned when considering a subspace of the task space (see our case study in appendix \ref{app:additional-results_parkour}). This might be due to the complexity of learning the climbing gait w.r.t walking or swimming, as it requires for instance good coordination skills between the arms and the grasping actions.
For the two other morphologies (bipedal walker and fish), results obtained are also low (respectively less than $60\%$ and $50\%$) and have a high variance (especially for the fish) considering that our test sets contain feasible tasks. This makes the Parkour environment an open challenge for future work on designing ACL algorithms.

\section{Open-Source release of \benchname}
With the open-source release of \benchname~ (version \textit{1.0}), we hope to provide a tool that can be used as a step towards thorough comparison and better understanding of current and future ACL methods. \benchname's documented repository features the code of our environments, embodiments, DRL students, as well as implementations of all ACL methods compared in this paper. All of these parts use APIs we provide such that one can easily add its ACL method, learning algorithm, and new embodiment or environment. We hope this will foster community-driven contributions to extend \benchname~in order to broaden its impact and adapt it to the future of ACL. We also provide the code we used to reproduce our experiments, as well as Jupyter notebooks allowing to generate all the figures showed in this paper. Finally, we release the results of our benchmark, allowing one to load them and compare its ACL method against baselines without having to reproduce our large-scale experiments.

\section{Discussion and Conclusion}

In this article we presented \benchname~\textit{1.0}, a first extensive testbed to design and compare ACL algorithms. It features unit-tests environments to assess the efficiency of a given teacher algorithm on multiple core skills and the Parkour environment, which provides a challenging teaching scenario that has yet to be solved. We used \benchname~ to conduct a comparative study of existing ACL algorithms. Throughout our experiments, we identified that 1) current ACL approaches not using expert knowledge matched and even outperformed (e.g. ALP-GMM) other approaches using high amounts of expert knowledge, and 2) the Parkour environment is far from solved, which makes it a good candidate as a testbed when designing new ACL approaches.

\paragraph{Limitations \& future work.} An obvious extension of this work is the addition of recent ACL approaches proposed during or after our experimental campaign \citep{acl-value-disagreement,jiang2020prioritized}.
So far, all studied ACL algorithms struggled to detect feasible task subspaces in Parkour, hinting that more research is needed to improve the "progress niche detection" ability of current teacher algorithms.

\benchname~currently only features environments with low-dimensional PCG systems. Designing new environments with higher-dimensional PCG, that might require to learn low dimensional representations on which to apply ACL algorithms, is an interesting avenue. Besides, our current list of environments only studies 2D locomotion tasks inspired by ALP-GMM's original paper \citep{portelas2019} as well as other works on Deep RL and 2D locomotion \citep{agentdesignha,bipwalkerSong18,NEURIPS2019_ha_weight_agnostic,poet,poet2}. While we put maximal effort in building a thorough and fair analysis of ACL methods, we believe extending \benchname~with other environments (e.g. ProcGen \cite{procgen}, robotic manipulation) would make the benchmark even more informative.

Additionally, extending the benchmark to consider environment-conditioned goal selection \citep{settersolver,AMIGO} -- i.e. where teachers have to observe the initial episode state to infer admissible goals -- is also worth investigating. \benchname~ provides a distribution of diverse learners. To this respect, it could also serve as a testbed for \textit{Meta ACL} \citep{meta-acl}, i.e. algorithms \textit{learning to learn to teach} across a sequence of students.


\subsubsection*{Acknowledgments}
This work was supported by Microsoft Research through its PhD Scholarship Programme. Experiments presented in this paper were carried out using 1) the PlaFRIM experimental testbed, supported by Inria, CNRS (LABRI and IMB), Université de Bordeaux, Bordeaux INP and Conseil Régional d’Aquitaine (see https://www.plafrim.fr/), 2) the computing facilities MCIA (Mésocentre de Calcul Intensif Aquitain) of the Université de Bordeaux and of the Université de Pau et des Pays de l'Adour, and 3) the HPC resources of IDRIS under the allocation 2020-[A0091011996] made by GENCI.

\bibliography{bibliography}
\bibliographystyle{icml2021}

\clearpage
\onecolumn
\input{appendices.tex}
\end{document}

%% file: math_commands.tex
\usepackage{amsmath,amsfonts,bm}

\def\eqref#1{equation~\ref{#1}}

\def\1{\bm{1}}

\DeclareMathAlphabet{\mathsfit}{\encodingdefault}{\sfdefault}{m}{sl}
\SetMathAlphabet{\mathsfit}{bold}{\encodingdefault}{\sfdefault}{bx}{n}



%% file: appendices.tex
\appendix
\onecolumn
\input{appendices/acl_details}

\clearpage
\input{appendices/env_details}

\clearpage
\input{appendices/expe_details}

\clearpage
\input{appendices/additional_results}


%% file: appendices/acl_details.tex
\section{Details on ACL baselines}
\label{app:acl-details}
In this section, we give details about our implementations of ACL methods, as well as their hyperparameters tuning. 

\subsection{Implementation details}
\paragraph{Random} We use as baseline a random teacher, which samples tasks using a uniform distribution over the task space.

\paragraph{ADR}  \citet{OpenAI2019SolvingRC} introduced \textit{Automatic Domain Randomization} (ADR), an ACL method relying on the idea of \textit{Domain Randomization} \citep{domain_randomization, sim_to_real_domain_rand}. Instead of sampling tasks over the whole task space, ADR starts from a distribution centered on a single example easy for the student and progressively grows the distribution according to the learning agent's performance. Using this mechanism, it increases the difficulty of the tasks proposed to the student while still sampling in previously seen regions in order to try reducing potential forgetting.

This sampling distribution $P_{\phi}$ is parameterized by $\phi \in \mathbb{R}^{2d}$ (with $d$ the number of dimensions of the task space). For each dimension, a lower and upper boundary are set $\phi = \{\phi_i^L, \phi_i^H\}_{i=1}^d$ allowing to sample uniformly on each dimension using these boundaries and obtain a task $\lambda$:
\[P_{\phi}(\lambda) = \prod_{i=1}^d U(\phi_i^L, \phi_i^H)\]

At the beginning, $\phi$ is centered on a single example (i.e.\ $\phi^L_i = \phi^H_i \; \forall i$). Then, at each episode, 1) ADR starts by sampling a new task $\lambda \sim P_{\phi}$. Following this, 2) ADR chooses with a probability $p_b$ whether to modify $\lambda$ in order to explore the task space or not. It thus samples a value $\epsilon$ uniformly in $[0; 1]$ and checks whether $\epsilon < p_b$. If this is not the case, ADR simply sends $\lambda$ to the environment.

Otherwise, 3) ADR selects uniformly one of the dimensions of the task space, which we will call $j$ as an example. Following this, 4) one of the two boundaries $\phi_j^L$ or $\phi_j^H$ is selected ($50\%$ chances for each boundary). Finally, 5) ADR replaces the $j$-th value of $\lambda$ by the selected boundary and sends $\lambda$ to the environment.

Moreover, ADR keeps a buffer $D^L_i$ and $D^H_i$ for each dimension $i$ in the task space. Every time $\epsilon$ is greater than $p_b$ and a value of $\lambda$ is replaced by one of the selected boundary, ADR stores the episodic reward obtained at the end of the episode in the buffer associated to the selected boundary (e.g. the episodic reward is stored in $D^L_k$ if the $k$-th value of lambda was replaced by $\phi^L_k$). 

Every time one of the buffers' size reaches $m$, the average $\overline{p}$ of episodic reward  stored is calculated. Then, $\overline{p}$ is compared to two thresholds $t_L$ and $t_H$ (being hyperparameters of ADR) in order to know whether the boundary associated to the buffer must be reduced or increased.

As an example, let's say that $D^L_k$'s size reached $m$, meaning that $\phi^L_k$ is the associated dimension (i.e. a $\lambda$ sampled got its $k$-th value replaced by $\phi^L_k$ $m$ times). Its average episodic reward $\overline{p}$ is calculated. It is first compared to $t_L$ and, if $\overline{p} < t_l$, $\phi^L_k$ is increased by $\Delta$ (as $\phi^L_k$ is a lower boundary, this means that the task space is reduced). Similarly, if $\overline{p} > t_l$, $\phi^L_k$ is decreased by $\Delta$ (expanding the task space).

If instead of $D^L_k$ we take $D^H_k$, our task space has to be expanded or reduced in the same way: if $\overline{p} < t_L$ then $\phi^H_k$ is reduced by $\Delta$ (as it is now an upper boundary of the task space) and if $\overline{p} > t_H$ then $\phi^H_k$ is increased by $\Delta$. Finally, note that whenever one buffer's size reaches $m$, it is then emptied.

As no implementation was provided by the authors, we propose here an implementation being as close as possible to the algorithms given in \citet{OpenAI2019SolvingRC}.

\paragraph{RIAC} Proposed in \citet{riac}, Robust Intelligent Adaptive Curiosity is based on the recursive splitting of the task space in hyperboxes, called regions. One region is split in two whenever a pre-defined number $max_s$ of sampled tasks originate from the region. The split value is chosen such that there is maximal Learning Progress (LP) difference between the two regions, while maintaining a size $min_d$ (i.e. a ratio of the size of the whole task space) for each region.  The number of possible split to attempt is parameterized by $n$. We reuse the implementation and the value of the hyperparameters not mentioned here from \citet{portelas2019}. RIAC does not require expert knowledge. 

\paragraph{Covar-GMM} Covar-GMM was proposed in \citet{moulinfriergmm}. As for RIAC, it does not require expert knowledge and is based on learning progress. The core idea of Covar-GMM is to fit a Gaussian Mixture Model (of maximum size $max_k$) every $n$ episodes on recently sampled tasks \textit{concatenated with both a time dimension and a competence dimension}. The Gaussian from which to sample a new task is then chosen proportionally to its respective learning progress, defined as the positive correlation between time and competence. Additionally, in order to preserve exploration, Covar-GMM has a probability $r_p$ of sampling a task uniformly random instead of using one of its Gaussians.  We use the implementation and hyperparameters from \citet{portelas2019} which uses Absolute Learning Progress (ALP) instead of LP.

Moreover, as aforementioned in section \ref{sec:experiments}, we modified the implementation to make it use expert knowledge (i.e. an initial distribution) when provided. Hence, instead of uniformly sampling tasks over the whole task space during the bootstrap phase at the beginning of training, Covar-GMM samples tasks from an initial Gaussian distribution of tasks provided by the expert. 

\paragraph{ALP-GMM} ALP-GMM is an ACL algorithm inspired from Covar-GMM, proposed in \citet{portelas2019}. Instead of relying on time competence correlation, which only allows to compute ALP over a single GMM fit, it computes a per-task ALP from the entire history of sampled tasks using a knn-based approach similar to those proposed in \citet{imgep}. Recent tasks are periodically used to fit a GMM on recently sampled tasks \textit{concatenated with their respective ALP value}. The Gaussian from which to sample is then selected based on its mean ALP dimension. ALP-GMM does not require expert knowledge and has the same hyperparameters as Covar-GMM. We reused the implementation and hyperparameters (except $max_k$, $n$ and $r_p$) provided by \citet{portelas2019}.

Additionally, as for Covar-GMM, we added the possibility to ALP-GMM to bootstrap tasks for an initial Gaussian distribution if the latter is provided, instead of uniformly bootstrapping tasks.

\paragraph{Goal-GAN}
Another teacher algorithm we included in this benchmark is called GoalGAN, and relies on the idea of sampling goals (i.e.\ states to reach in the environment) where the agent performs neither too well nor to badly, called \textit{Goals Of Intermediate Difficulty} (GOID). However, as this goal generation introduces a curriculum in the agent's learning, one can see the goal selection process as a task selection process. We will thus call them tasks instead of goals in the following description. For sampling, \citet{goalgan} proposed to use a modified version of a \textit{Generative Adversarial Network} (GAN) \citep{goodfellow_generative_2014} where the generator network is used to generate tasks for the student given a random noise, and the discriminator is trained to classify whether these tasks are of "intermediate difficulty". To define such an "intermediate difficulty", GoalGAN uses a binary reward signal defining whether the student succeeded in the proposed task. As our environments return scalar rewards, this implies a function interpreter hand-designed by an expert (in our case we set a threshold on the scalar reward, as explained in appendix \ref{app:expe-details}). For each task sampled, the teacher proposes it multiple times ($n_{rollouts}$) to the student and then calculates the average of successes obtained (lying in $[0; 1]$). Using a lower threshold $R_{min}$ and an upper threshold $R_{max}$, GoalGAN calculates if the average lies in this interval of tasks neither too easy (with an average of successes very high) nor too hard (with an average of successes very low). If this is the case, this task is labelled as $1$ for the discriminator ($0$ otherwise). This new task is then stored in a buffer (except if it already exists in the buffer a task at an euclidean distance smaller than $\epsilon$ from our new task). Every time a task has to be sampled, in order to prevent the GAN from forgetting previously seen GOIDs, the algorithm has the probability $p_{old}$ of uniformly sampling from the buffer instead of using the GAN. Finally, the GAN is trained using the tasks previously sampled every $n$ episodes.

Note that, in order to help the GAN to generate tasks in a feasible subspace of the task space at the beginning of training, GoalGAN can also pretrain its GAN using trivial tasks. In the original paper, as tasks are states, authors proposed to use the student to interact with the environment for a few steps, and use collected states as achievable tasks. However, in our case, this is not possible. We thus chose to reuse the same trick as the one in \citep{selfpaceddrl}, that uses an initial Gaussian distribution to sample tasks and label them as positives (i.e.\ tasks of intermediate difficulty) in order to pretrain the GAN with them. See appendix \ref{app:expe-details} for the way we designed this initial distribution.

We reused and wrapped the version\footnote{\url{https://github.com/psclklnk/spdl}} of GoalGAN implemented by \citet{selfpaceddrl}, which is a slightly modified implementation of the original one made by \citet{goalgan}. Our generator network takes an input that has the same number of dimensions as our task space, and uses two layers of $256$ neurons with ReLU activation (and TanH activation for the last layer). Our discriminator uses two layers of $128$ neurons. For $\epsilon$, we used a distance of $10\%$ on each dimension of the task space. As per \citet{goalgan}, we set $R_{min}$ to $0.25$ and $R_{max}$ to $0.75$. Finally, as in the implementation made by \citet{selfpaceddrl}, we set the amount of noise $\delta$ added to each goal sampled by the generator network as a proportion of the size of the task space.

\paragraph{Self-Paced}
Proposed by \citet{selfpaceddrl}, \textit{Self-Paced Deep Reinforcement Learning} (SPDL) samples tasks from a distribution that progressively moves towards a target distribution. The intuition behind it can be seen as similar to the one behind ADR, as the idea is to start from an initial task space and progressively shift it towards a target space, while adapting the pace to the agent's performance. However here, all task distributions (initial, current and target) are Gaussian distributions. SPDL thus maintains a current task distribution from which it samples tasks and changes it over training. This distribution shift is seen as an optimization problem using a dual objective maximizing the agent's performance over the current task space, while minimizing the Kullback-Leibler (KL) divergence between the current task distribution and the target task distribution. This forces the task selection function to propose tasks where the agent performs well while progressively going towards the target task space. 

Initially designed for non-episodic RL setups, SPDL, unlike all our other teachers, receives information at every step of the student in the environment. After an offset of $n_{\mathrm{OFFSET}}$ first steps, and then every $n_{\mathrm{STEP}}$ steps, the algorithm estimates the expected return for the task sampled $\mathrm{E}_{p(c)}[J(\pi, c)]$ using the value estimator function of the current student (with $p(c)$ the current task distribution, $\pi$ the current policy of the student, and $J(\pi, c)$ the expected return for the task $c$ with policy $\pi$). 

With this, SPDL updates its current sampling distribution in order to maximize the following objective w.r.t. the current task distribution $p(c)$:
\[\max_{p(c)} \mathrm{E}_{p(c)}[J(\pi, c)]\]

Additionally, a penalty term is added to this objective function, such that the KL divergence between $p(c)$ and the target distribution $\mu(c)$ is minimized. This penalty term is controlled by an $\alpha$ parameter automatically adjusted. This parameter is first set to $0$ for $K_{\alpha}$ optimization steps and is then adjusted in order to maintain a constant proportion $\zeta$ between the KL divergence penalty and the expected reward term (see \citet{selfpaceddrl} for more details on the way $\alpha$ is calculated). This optimization step is made such that the shift of distribution is not bigger than $\epsilon$ (i.e. $s.t. D_{\mathrm{KL}}(p(c)||q(c)) \leq \epsilon$ with a shift from $p(c)$ to $q(c)$).

We reused the same implementation made by \citet{selfpaceddrl} and wrapped it to our teacher architecture. However, as shown in section \ref{sec:experiments}, using a Gaussian target distribution does not match with our Stump Tracks test set where tasks are uniformly sampled over the whole task space. In order to solve this issue, some adaptations to its architecture could be explored (e.g. using a truncated Gaussian as target distribution to get closer to a uniform distribution). While not provided yet in \benchname, we are currently working along with SPDL's authors on these modifications in order to show a fairer comparison of this promising method.

For the value estimators, we used the value network of both our PPO and SAC implementations (with the value network sharing its weights with the policy network for PPO). For the calculation of $\alpha$, we chose to use the average reward, as in the experiments of \citet{selfpaceddrl}. We did not use the lower bound restriction on the standard deviation of the task distribution $\sigma_{\mathrm{LB}}$ proposed in \citet{selfpaceddrl} as our target distributions were very large (see appendix \ref{app:expe-details}).

\paragraph{Setter-Solver} Finally, the last ACL algorithm we implemented here is Setter-Solver \citep{settersolver}. In a very similar way to Goal-GAN, this method uses two neural networks: a \textit{Judge} (replacing the discriminator) and a \textit{Setter} (replacing the generator) outputting a task given a feasibility scalar in $[0; 1]$. During the training, the \textit{Judge} is trained to output the right feasibility given a task sampled, and is used in the \textit{Setter}'s losses to encourage the latter to sample tasks where the predicted feasibility was close to the real one. The \textit{Setter} is also trained to sample tasks the student has succeeded (i.e.\ using a binary reward signal as Goal-GAN) while maximizing an entropy criterion encouraging it to sample diverse tasks.

For the implementation, \citet{settersolver} provided code to help reproducibility that implements both the \textit{Setter} and \textit{Judge}, but did not include neither losses nor optimization functions. Therefore, we provide here our own implementation of the full Setter-Solver algorithm trying to be as close as possible to the paper's details. We reused the code provided for the two neural networks and modified it to add losses, optimizers, and some modifications to better integrate it to our architecture. We kept the tricks added in the code provided by authors that uses a non-zero uniform function to sample the feasibility and a clipped sigmoid in the \textit{Setter}'s output. Concerning the generator network, we kept the hyperparameters of the paper (i.e.\ a RNVP \citep{RNVP} with three blocks of three layers) except the size of hidden layers $n_{\mathrm{HIDDEN}}$ that we optimized. We also reused the three layers of $64$ neurons architecture for the \textit{Judge} as per the paper. Note that we used an Adam optimizer with a learning rate of $3\cdot 10^{-4}$ for both the \textit{Setter} and the \textit{Judge}, while this was not precised for the \textit{Judge} in \citet{settersolver}. We optimized the upper bound $\delta$ of the uniformly sampled noise that is added to succeeded tasks in the validity \textit{Setter}'s loss, as well as the update frequency $n$.

We did not use the conditioned version of the \textit{Setter} or \textit{Judge}. Indeed, first we generate the task before obtaining the first observation in our case as opposed to \citet{settersolver}, and also because the first observation of an embodiment is always the same as both our environments have a startpad (see appendix \ref{app:env-details}). Finally, we did not use the additional target distribution (called \textit{desired goal distribution} in the original paper) loss that use a Wassertein discriminator \citep{arjovsky_wasserstein_2017} to predict whether a task predicted belongs to the target distribution. Indeed, as shown in \citet{settersolver}, using the targeted version of Setter-Solver offers more sample efficiency but leads to similar final results. Moreover, in our case,  a target distribution is known only in the \textit{High expert knowledge} setup of the challenge-specific experiments, in addition of having this part not implemented at all in the code provided by authors. We thus leave this upgrade to future work.

\subsection{Hyperparameters tuning}
In order to tune the different ACL methods to our experiments, we chose to perform a grid-search using our Stump Tracks environment with its original task space. As the Parkour is partly extended from it, in addition of the challenge-specific experiments, this environment offered us an appropriate setup. Each point sampled in the grid-search was trained for $7$ million steps (instead of the $20$ millions used in our experiments) with $16$ seeds in order to reduce the (already high) computational cost. At the end of training, we calculated the percentage of mastered tasks on test set for each seed. The combination of hyperparameters having the best average over its seeds was chosen as the configuration for the benchmark.

In order to make the grid-search as fair as possible between the different ACL methods, given that the number of hyperparameters differs from one method to another, we sampled the same number of points for each teacher: $70$ ($\pm 10$). The hyperparameters to tune for each teacher, as well as their values, were chosen following the recommendations given by their original paper.

Moreover, we chose to tune the teachers in what we call their “original” expert knowledge version (i.e. they have access to the same amount of prior knowledge as the one they used in their paper). Hence, teachers requiring expert knowledge use our high expert knowledge setup, and algorithms such as ALP-GMM use no expert knowledge.

Table \ref{table:hp_tuning} shows the values we tested for each hyperparameter and the combinations that obtained the best result.

\begin{table}[H]
\caption{Hyperparameters tuning of the ACL methods.}
\label{table:hp_tuning}
\vskip 0.15in
\setlength\tabcolsep{1.4pt}
\begin{center}
\begin{small}
\begin{sc}
    \begin{tabular}{|c|c|c|c|}
        \hline
        ACL method & Hyperparameter & Possible values & Best value\\ \hline\hline
        \textbf{ADR}           & $t_L$ & $[0, 50]$ &    $0$ \\ \hline
        ADR                    & $t_H$ & $[180, 230, 280]$ &    $180$ \\ \hline
        ADR                    & $p_b$ & $[0.3, 0.5, 0.7]$ &    $0.7$ \\ \hline
        ADR                    & $m$ & $[10, 20]$ &    $10$ \\ \hline
        ADR                    & $\Delta$ & $[0.05, 0.1]$ &    $0.1$ \\ \hline\hline
        \textbf{RIAC}          & $max_s$ & $[50, 150, 250, 350]$ &    $150$ \\ \hline
        RIAC                   & $n$ & $[25, 50, 75, 100]$ &    $75$ \\ \hline
        RIAC                   & $min_d$ & $[0.0677, 0.1, 0.1677, 0.2]$ &    $0.1$ \\ \hline\hline
        \textbf{Covar-GMM}     & $n$ & $[50, 150, 250, 350]$ &    $150$ \\ \hline
        Covar-GMM              & $max_k$ & $[5, 10, 15, 20]$ &    $15$ \\ \hline
        Covar-GMM              & $r_p$ & $[0.05, 0.1, 0.2, 0.3]$ &    $0.1$ \\ \hline\hline
        \textbf{ALP-GMM}       & $n$ & $[50, 150, 250, 350]$ &    $150$ \\ \hline
        ALP-GMM                & $max_k$ & $[5, 10, 15, 20]$ &    $10$ \\ \hline
        ALP-GMM                & $r_p$ & $[0.05, 0.1, 0.2, 0.3]$ &    $0.05$ \\ \hline\hline
        \textbf{GoalGAN}       & $\delta$ & $[0.01, 0.05, 0.1]$ &    $0.01$ \\ \hline
        GoalGAN                & $n$ & $[100, 200, 300]$ &    $100$ \\ \hline
        GoalGAN                & $p_{old}$ & $[0.1, 0.2, 03]$ &    $0.2$ \\ \hline
        GoalGAN                & $n_{rollouts}$ & $[2, 5, 10]$ &    $2$ \\ \hline\hline
        \textbf{SPDL}          & $n_{\mathrm{OFFSET}}$ & $[100000, 200000]$ &    $200000$ \\ \hline
        SPDL                   & $n_{\mathrm{STEP}}$ & $[50000, 100000]$ &    $100000$ \\ \hline
        SPDL                   & $K_{\alpha}$ & $[0, 5, 10]$ &    $0$ \\ \hline
        SPDL                   & $\zeta$ & $[0.05, 0.25, 0.5]$ &    $0.05$ \\ \hline
        SPDL                   & $\epsilon$ & $[0.1, 0.8]$ &    $0.8$ \\ \hline\hline
        \textbf{Setter-Solver} & $n$ & $[50, 100, 200, 300]$ &    $100$ \\ \hline
        Setter-Solver          & $\delta$ & $[0.005, 0.01, 0.05, 0.1]$ &    $0.05$ \\ \hline
        Setter-Solver          & $n_{\mathrm{HIDDEN}}$ & $[64, 128, 256, 512]$ &    $128$ \\ \hline
    \end{tabular}
\end{sc}
\end{small}
\end{center}
\vskip -0.1in
\end{table}

%% file: appendices/env_details.tex
\section{Environment details}\label{app:env-details}
In this section, we give details about our two environments, their PCG algorithm, as well as some analysis about their task space. Note that our two environments follow the OpenAI Gym's interface and provide after each step, in addition of usual information (observation, reward, and whether the episode terminated), a binary value set to $1$ if the cumulative reward of the episode reached $230$. Additionally, we provide extra information and videos of our environments and embodiments, as well as policies learned at at \url{http://developmentalsystems.org/TeachMyAgent/}.

\subsection{Stump Tracks}
We present the Stump Tracks environment, an extended version of the environment introduced by \citet{portelas2019}. We only use two of the initially introduced dimensions of the procedural generation of task: stumps' height $\mu_s$ and spacing $\Delta_s$s. As in \citet{portelas2019}, $\mu_s$ is used as the mean of a Gaussian distribution with standard deviation $0.1$. Each stump has thus its height sampled from this Gaussian distribution and is placed at distance $\Delta_s$ from the previous one. We bound differently this task space depending on the experiment we perform, as explained in appendix \ref{app:expe-details}.

We kept the same observation space with $10$ values indicating distance of the next object detected by lidars, head angle and velocity (linear and angular), as well as information from the embodiment (angle and speed of joints and also whether the lower limbs have contact with the ground). For information concerning the embodiment, the size of observation depends on the embodiment, as the number of joints varies (see below in \ref{app:env-details_bestiary}). We also kept the action space controlling joints with a torque.

\subsection{Parkour}
We introduce the Parkour, a Box2D parkour track inspired from the Stump Tracks and the environment introduced in \citet{poet2}. It features different milieu in a complex task space. 

\subsubsection{Procedural generation}
\paragraph{CPPN-encoded terrain} First, similarly to the Stump Tracks, our Parkour features a ground (that has the same length as the one in Stump Tracks) where the agent starts at the leftmost side and has to reach the rightmost side. However, this ground is no longer flat and rather, as in \citet{poet2}, generated using a function outputted by a neural network called CPPN \citep{CPPN}. This network takes in input a $x$ position and outputs the associated $y$ position of the ground. Using this, one can slide the CPPN over the possible $x$ positions of the track in order to obtain the terrain. This method has the advantage of being able to easily generate non-linear and very diverse terrains as shown in \citet{poet2}, while being light and fast to use as this only needs inference from the network. While CPPNs are usually used in an evolutionary setup where the architecture and weights are mutated, we chose here to rather initialize an arbitrary architecture and random weights and keep them fixed. For this architecture, we chose to use a four layers feedforward neural network with $64$ units per layer and an alternation of TanH and Softplus activations (except for the output head which uses a linear activation) inspired from \citet{ha2016abstract}. Weights were sampled from a Gaussian distribution with mean $0$ and standard deviation of $1$. In addition of its $x$ input, we added to our network three inputs that are set before generating the terrain as parameters controlling the generation. This vector $\theta$ of size $3$ acts in a similar way as noise vector does in GANs for instance. Its size was chosen such that it allows to analyse the generation space and maintain the overall task space's number of dimensions quite small. As for the parameters in Stump Tracks, we bounded the space of values an ACL method could sample in $\theta$. For this, we provide three hand-designed setups (easy, medium and hard) differing from the size of the resulting task space and the amount of feasible tasks in it (see appendix \ref{app:expe-details-parkour}). 

Moreover, in addition of the $y$ output of the ground, we added another output head $y_c$ in order to create a ceiling in our tracks. As in Stump Tracks, the terrain starts with a flat startpad region (with a fixed distance between the ground and the ceiling) where the agent appears. Once $Y = (y_i)_{i \in X}$ and $Y_c = (y_{c_i})_{i \in X}$ generated by the CPPN, with $X$ all the possible $x$ positions in the track, we align them to their respective startpad:
\[y_i = y_i + startpad_g - y_0 \; \forall i \in Y\]
\[y_{c_i} = y_{c_i} + startpad_c - y_{c_0} \; \forall i \in Y_c\]
with $startpad_g$, $startpad_c$ being respectively the $y$ position of the ground startpad and ceiling startpad, and $y_0$, $y_{c_0}$ respectively the first $y$ position of the ground and the ceiling outputted by our CPPN.

\begin{figure}[H]
\vskip 0.2in
\begin{center}
\centerline{\includegraphics[width=0.7\textwidth]{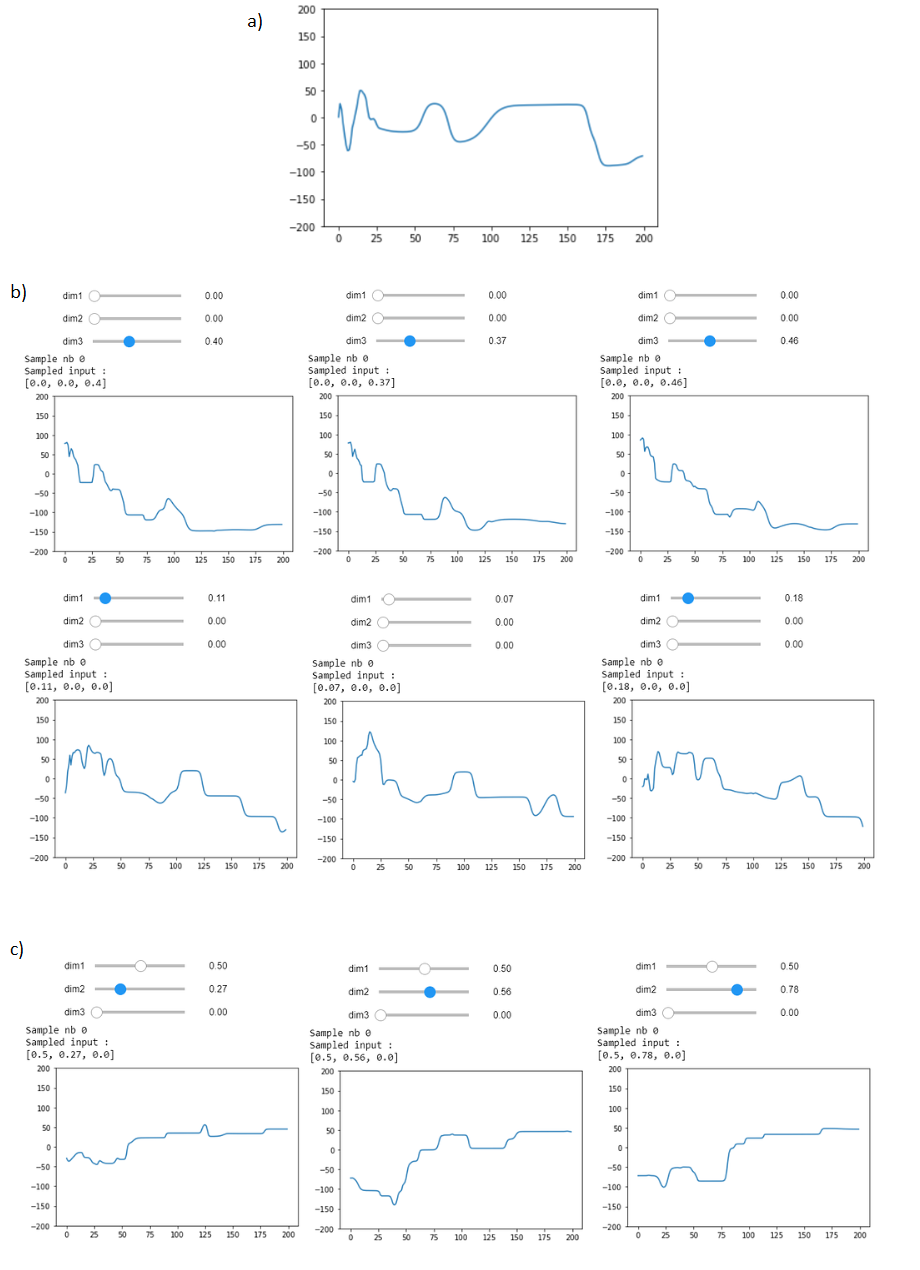}}
\caption{Overview of the input space of $\theta$. First, in a) one can see the function generated when all the values of the input vector are set to zero. Secondly, in b) we can see that small changes over the space lead to similar functions and that big changes lead to very different results, showing that local similarity is maintained over the task space. Finally, c) shows how the difficulty landscape of $\theta$ can be rugged, as moving along the second dimension leads to terrains having a very different difficulty level.}
\label{fig:CPPN_Input_Space_Analysis}
\end{center}
\vskip -0.2in
\end{figure}

\begin{figure}[H]
\vskip 0.2in
\begin{center}
\centerline{\includegraphics[width=0.9\textwidth]{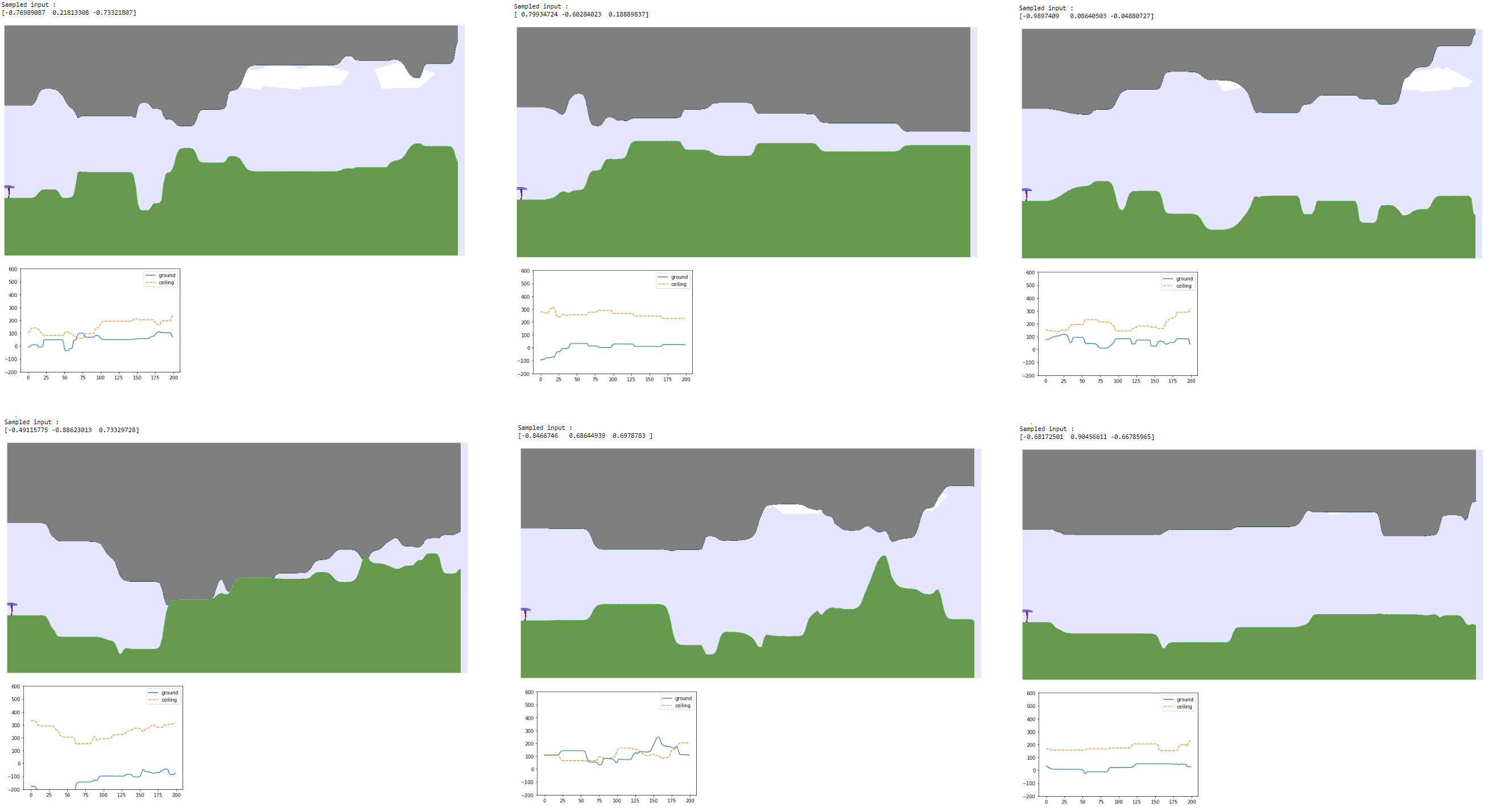}}
\caption{Here are some examples of generated tasks in the Parkour environment. While most of them seem too hard for a classic bipedal walker, the bottom left task is a good example of an unfeasible task, no matter which embodiment is used.}
\label{fig:Ground_Ceiling_Examples}
\end{center}
\vskip -0.2in
\end{figure}

Using this non-linear generator (i.e. CPPN) allows us to have an input space where the difficulty landscape of the task space is rugged. Indeed, in addition of generating two non-linear functions for our ground and ceiling, the two latter can cross each other, creating unfeasible tasks (see figure \ref{fig:Ground_Ceiling_Examples}). Additionally, our CPPN also makes the definition of prior knowledge over the input space more complex, as shown in figure \ref{fig:CPPN_Input_Space_Analysis}.

Finally, as shown in figure \ref{fig:Ground_Ceiling_Examples}, we smoothed the values of $Y$ and $Y_c$ by a parameter $\delta$ ($=10$ in the training distribution) in order to make the roughness of the terrains adapted to our embodiments.

\paragraph{Creepers} Once the terrain generated, we add what we call "creepers". Similarly to the stumps, we create objects at distance $\Delta_c$ from one another and of height sampled using a Gaussian distribution of mean $\mu_c$ and standard deviation $0.1$ (the width can also be controlled but was fixed to $0.25$ in our experiments). However, creepers are not obstacles for agents as stumps but rather graspable objects that embodiments can go through. Moreover, even though not used in our experiments, we provide the possibility to make creepers more realistic by dividing every creeper in multiple rectangles of height at most $1$ linked with a rotating joint. As shown on our website , this creates creepers on which the climbers can swing.

\paragraph{Water} Finally, we added a last dimension to our task space controlling the "water" level. Water is simulated using a rectangle object that the agent can go through and in which physics change (see below). This rectangle's width equals the terrain's width and its height is controlled by a parameter $\tau \in [0;1]$ with $0$ being an arbitrary lower limit the ground can reach and $1$ the highest point of the current ceiling (generated by the CPPN for the current task).

\subsubsection{Physics}
As previously mentioned, we introduced creepers and water along with new physics. First, in order to make our creepers graspable by the agents, we added sensors to the end of limb of certain embodiments (see section \ref{app:env-details_bestiary} below). Every time one of these sensors enters in contact with a creeper, we look at the action in the action space of the agent that is associated to this sensor. If its value is greater than $0$, we create a rotational joint between the sensor and the creeper at the contact point. As long as this action is greater than $0$, the joint remains. As soon as the action goes negative or equals $0$, we delete the joint (releasing the agent's limb from the creeper) and start watching again for contact. Note that, in order to better see whether the agent grasps a creeper, we color its sensors in red when a joint exists and in yellow otherwise (see our website). Additionally, in order to help the learning agent, we also make the ceiling graspable.

Secondly, concerning the water, we simulated a buoyancy force (inspired from \citet{buoyancy}) when an object enters water given its density compared to the water's density (set to $1$). In addition, we implemented a "drag" and a "lift" force that simulate the resistance applied when an object moves in water and slows down the movement. Finally, we added a "push" force applied to an object having an angular velocity. This force simulates the fact that applying a torque to a rotational joint makes the object attached to the joint "push" the water and move (i.e. have a linear force applied). With these forces, we were able to simulate in a simplified way some physics of water, which resulted in very natural policies learned from our agents (see our website).

Finally, we simulated the fact that each embodiment is suited for one (or several) milieu, creating types of agents. Indeed, we first consider swimming agents that die (i.e. the actions sent by the DRL student to the environment no longer have effects on the motors of the embodiment) after spending more than $600$ consecutive steps outside water. On the contrary, the two other types named climbers and walkers cannot survive underwater more than $600$ consecutive steps. 
Both walkers and swimmers are allowed to have collisions with their body (including their head in the Parkour), whereas climbers are not allowed to touch the ground with any part of their body. Note that, while walkers appear with their legs touching the ground, swimmers appear a bit above the ground and climbers appear with all of their sensors attached to the ceiling (see figure \ref{fig:bench-vizu}). 

All of these physics introduce the fact that an ACL teacher has to propose tasks in the right milieu for the current embodiment (i.e. mostly underwater for swimmers so that they do not die, with creepers and a ceiling high enough for climbers so that they do not touch the ground or die in water and with no water for walker so that they do not drown) in order to make the student learn.  

\subsubsection{Observation and action space}
As in Stump Tracks, the agent is rewarded for moving forward and penalized for torque usage. An episode lasts $2000$ steps unless the agent reaches the end of the track before or if a part of its body touches the ground if the embodiment is a climber. We also reused the $10$ lidars per agent that were used in the Stump Tracks with all the lidars starting from the center of the head of the morphology. However, we modified them such that three configurations of covering exist (see the three tasks shown in figure \ref{fig:bench-vizu}):
\begin{itemize}
    \item $90\degree$ from below the agent to ahead of it (used by walkers, as in Stump Tracks)
    \item $180\degree$ from below the agent to above it (used by swimmers)
    \item $90\degree$ from ahead of the agent to above it (used by climbers)
\end{itemize}

Moreover, in addition of the distance to the next object detected by each lidar, we added an information concerning the type of object detected by the lidar ($-1$ if water, $1$ if creeper, $0$ otherwise) such that the agent knows whether the object detected is an obstacle or can be passed through. Note also that once their origin point overlaps an object (e.g. water), lidars no longer detect it. Hence the lidars of an agent underwater no longer detect water (which would have made lidars useless as they would have only detected water). Therefore, in order to inform the DRL student whether the embodiment is underwater, we added an observation that is set to $1$ if the agent's head is under the water level and $0$ otherwise. Similarly, we added a binary observation telling whether the agent is dead or not (i.e. the actions we send to its motors no longer have impact). In addition, we kept the same information concerning the agent's head as in Stump Tracks (angle, linear velocity and angular velocity) as well as observations for each motor (angle and speed of joint as well as contact information for some of the attached limb). Finally, we added two binary observations per sensor (if the agent has sensors) telling whether the sensor has contact with a graspable surface and whether it is already attached with a joint. Without considering the information about motors and sensors which depend on the morphology, all of the information listed above create an observation vector of size $26$. Note that, additionally, we provide the information to the teacher at each step whether the cumulative reward of the episode has reached $230$ for the users using a binary reward.

Finally, for the action space, we kept the same behaviour as the one used in Stump Tracks (i.e. each agent has motors which are controlled through a torque value in $[-1;1]$). Moreover, we added an action in $[-1;1]$ per sensor for climbers to say whether this sensor must grasp (if it has contact with a graspable surface) or release. 

\subsection{Morphologies} \label{app:env-details_bestiary}
We included in our benchmark the classic bipedal walker as well as its two modified versions  introduced in \citet{portelas2019}: the short bipedal and the quadrupedal. For these three agents, we kept in their implementation the additional penalty for having an angle different than zero on their head, which was already in \citet{portelas2019}. Additionally, we created new walkers such as the spider or the millipede shown in figure \ref{fig:bench-vizu}. See our repository and website for the exhaustive list of embodiments we provide.

We introduce another type of morphologies: climbers. We propose two agents: a chimpanzee-like embodiment, as well as its simplified version without legs (reducing the action space to simplify the learning task). These two agents have two arms with two sensors at their extremity allowing them to grasp creepers or the ceiling. 

Both walkers and climbers have a density of $1$ on their legs and arms, and a density of $5$ on their body and head, making them simply "sink" in water.

Finally, we created swimming morphologies with each of their body part having the same density as the water, making them in a zero-gravity setting when fully underwater. We propose a fish-like embodiment (see figure \ref{fig:bench-vizu}) with a fin and a tale that can wave its body to move (as well as moving its fin).

Note that we also included an amphibious bipedal walker allowed to survive both underwater and outside water. This gave interesting swimming policies as shown on our website ( \url{http://developmentalsystems.org/TeachMyAgent/}).

%% file: appendices/expe_details.tex
\section{Experimental details}\label{app:expe-details}
In this section, we give details about the setups of our experiments.

\subsection{DRL Students}
We used the 0.1.1 version of OpenAI Spinningup's implementation of SAC that uses Tensorflow, as in \citet{portelas2019}. We modified it such that a teacher could set a task at each reset of the environment. We also kept the same hyperparameters as the ones used in \citet{portelas2019}:
\begin{itemize}
    \item A two layers feedforward network with $400$/$300$ units per hidden layer (ReLU activation) for both the value and policy network (using TanH activation on the output layer for the latter)
    \item An entropy coefficient of $0.005$
    \item A learning rate of $0.001$
    \item A mini-batch update every $10$ steps using $1000$ randomly sampled experiences from a buffer of size $2$ millions
\end{itemize}

For PPO, we used OpenAI Baselines' (Tensorflow) implementation. We used the same two layers neural network as in SAC for the policy and value networks (which share weights). We modified the runner sampling trajectories from the environment in order to use a single synchronous runner instead of multiple asynchronous ones. We used the environment's wrappers proposed in the OpeanAI Baselines' implementation to clip the actions and normalize the observations and rewards. We added a test environment (as well as a teacher that sets tasks) to test the agent's performance every $500000$ steps (as done with SAC). We also normalize the observations and rewards in this test environment using the same running average as the one used in the training environment, so that agent does not receive different information from both environments. We send to the teacher and monitor the original values of reward and observation sent by the environment instead of normalized ones.
We set the $\lambda$ factor of the Generalized Advantage Estimator to $0.95$, the clipping parameter $\epsilon$ to $0.2$ and the gradient clipping parameter to $0.5$.
Finally, we tuned the following hyperparameters using a grid-search on Stump Tracks for $10$ millions steps with stumps' height and spacing respectively in $[0;3]$ and $[0;6]$:
\begin{itemize}
    \item Size of experiences sampled between two updates: $2000$
    \item Number of epochs per update: $5$
    \item Learning rate: $0.0003$
    \item Batch size: $1000$
    \item Value function coefficient in loss: $0.5$
    \item Entropy coefficient in loss: $0.0$
\end{itemize}

Note that for both our DRL students, we used $\gamma = 0.99$.

\subsection{General experimental setup}
We call an experiment the repetition, using different seeds, of the training of a DRL student for $20$ millions steps using tasks chosen at every reset of the environment by a selected ACL teacher. The seed is used to initialize the state of random generators used in the teacher, DRL student and environment. We provide to the teacher the bounds (i.e. a $min$ and $max$ value for each dimension) of the task space before starting the experiment. The DRL student then interacts with the environment and asks the ACL teacher to set the task (i.e. a vector controlling the procedural generation) at every reset of the environment. Once the episode ended, the teacher receives either the cumulative reward or a binary reward (set to $1$ if the episodic reward is grater than $230$) for GoalGAN and Setter-Solver. Teachers like SPDL can additionally access to the information sent by the environment at every step, allowing non-episodic ACL methods to run in our testbed. 

Every $500000$ steps of the DRL student in the environment, we test its performance on $100$ predefined tasks (that we call test set). We monitor the episodic reward obtained on each of these tasks. We also monitor the average episodic reward obtained on the tasks seen by the student during the last $500000$ steps. We ask the teacher to sample $100$ tasks every $250000$ steps of the DRL student and store these tasks to monitor the evolution of the generated curriculum (see at \url{http://developmentalsystems.org/TeachMyAgent/}). For this sampling, we use the non-exploratory part of our teachers (e.g. ALP-GMM always samples from its GMM or ADR never sets one value to one of its bounds) and do not append these monitoring tasks to the buffers used by some teachers to avoid perturbing the teacher's process.

In our experiments we were able to run 8 seeds in parallel on a single Nvidia Tesla V100 GPU. In this setup, evaluating one ACL method requires approximately (based on ALP-GMM’s wall-clocktime):
\begin{itemize}
    \item 4608 gpu hours for all skill-specific experiments with 32 seeds.
    \item 168 gpu hours for the 48 seeded Parkour experiment.
\end{itemize}
Running both experiments would require 4776 gpu hours, or 48 hours on 100 Nvidia Tesla V100 GPUs. Users with smaller compute budgets could reduce the number of seeds (e.g. divide by 3) without strong statistical repercussions.

\subsection{Stump Tracks variants}
We used the Stump Tracks environment to create our challenge-specific comparison of the different ACL methods. We leveraged its two dimensional task space (stumps' height and spacing) to create experiments highlighting each of the $6$ challenges listed in section \ref{sec:intro}. Each experiment used $32$ seeds.

\subsubsection{Test sets}
We used the same test set in all our experiments on Stump Tracks in order to have common test setup to compare and analyse the performance of our different ACL methods. This test set is the same as the one used in \citet{portelas2019} with $100$ tasks evenly distributed over a task space with $\mu_s \in [0;3]$ and $\Delta_s \in [0;6]$. 

\subsubsection{Experiments}
In the following paragraphs, we detail the setup of each of our experiments used in the challenge-specific comparison.

\paragraph{Expert knowledge setups}
We allow three different amounts of prior knowledge about the task to our ACL teachers:
\begin{itemize}
    \item \textit{No expert knowledge}
    \item \textit{Low expert knowledge}
    \item \textit{High expert knowledge}
\end{itemize}

First, in the \textit{No expert knowledge} setup, no prior knowledge concerning the task is accessible. Hence, no reward mastery range (ADR, GOoalGAN and Setter-Solver) is given. Additionally, no prior knowledge concerning the task space like regions containing trivial tasks for the agent (e.g. for ADR or SPDL's initial distribution) or subspace containing the test tasks (e.g. for SPDL's target distribution) are known. However, we still provide these two distribution using the following method:
\begin{itemize}
    \item \textit{Initial distribution}: we sample the mean $\mu_{INITIAL}$ of a Gaussian distribution uniformly random over the task space. We choose the variance of each dimension such that the standard deviation over this dimension equals $10\%$ of the range of the dimension (as done when expert knowledge is accessible).
    \item \textit{Target distribution}: we provide a Gaussian distribution whose mean is set to the center of each dimension and standard deviation to one fourth of the range of each dimension (leading to more than $95\%$ of the samples that lie between the min and max of each dimension). This choice of target distribution was made to get closer to our true test distribution (uniform over the whole task space), while maintaining most of the sampled tasks inside our bounds. However, it is clear that this target distribution is not close enough to our test distribution to make SPDL proposing a good curriculum and lead to an agent learning an efficient policy to perform well in our test set.  As mentioned in section \ref{sec:experiments} and appendix \ref{app:acl-details}, using a Gaussian target distribution is not suited to our setup and would require modifications to make the target distribution match our true test distribution.
\end{itemize}
Hence in this setup, only ALP-GMM, RIAC, Covar-GMM and SPDL (even though its initial and target distribution do not give insightful prior knowledge) can run.

In the \textit{Low expert knowledge} setup, we give access to reward mastery range. Therefore, GoalGAN, Setter-Solver and ADR can now enter in the comparison. The initial distribution is still randomly sampled as explained above. It is used by GoalGAN to pretrain its GAN at the beginning of the training process, but also by ADR which starts with a single example being $\mu_{INITIAL}$.

Finally, for the \textit{High expert knowledge} setup, we give access to the information about regions of the task space. While the standard deviation of the initial distribution is still calculated in the same way (i.e. $10\%$ of the range of each dimension), we set $\mu_{INITIAL}$ to $[0; 6]$, with the values being respectively $\mu_s$ and $\Delta_s$. Hence, ADR now uses the task $[0; 6]$ as its initial task and GoalGAN pretrains its GAN with this distribution containing trivial tasks for the walking agent (as stumps are very small with a large spacing between them). SPDL also uses this new initial distribution, but keeps the same target distribution as we could not provide any distribution matching our real test distribution (i.e. uniform).
Note that, as mentioned in appendix \ref{app:acl-details}, ALP-GMM and Covar-GMM use this initial distribution in their bootstrap phase in this setup.

\paragraph{Mostly unfeasible task space}
In this experiment, we use SAC with a classic bipedal walker. We consider stumps with height greater than $3$ impossible to pass for a classic bipedal walker. Hence, in order to make most of the tasks in the task space unfeasible, we use in this experiment $\mu_s \in [0; 9]$ (and do not change $\Delta_s \in [0; 6]$) such that almost $80\%$ of the tasks are unfeasible.

\paragraph{Mostly trivial task space}
Similarly, we use in this experiment $\mu_s \in [-3; 3]$ (the Stump Tracks environments clips the negative values with $\mu_s = \max(0, \mu_s)$). Hence $50\%$ of the tasks in the task space will result in a Gaussian distribution used to generate stumps' height with mean $0$. We also use SAC with a classic bipedal walker.

\paragraph{Forgetting students}
We simulate the catastrophic forgetting behaviour by resetting all the variables of the computational graph of our DRL student (SAC here) as well as its buffers every $7$ millions steps (hence twice in a training of $20$ millions steps). All variables (e.g. weights, optimizer's variables...) are reinitialized the same way they were before starting the training and the experience buffer used by SAC is emptied. Note that we also use the classic bipedal walker as embodiment and did not modify the initial task space ($\mu_s \in [0;3]$ and $\Delta_s \in [0;6]$).

\paragraph{Rugged difficulty landscape}
In order to create a rugged difficulty landscape over our task space, we cut it into $4$ regions of same size and shuffle them (see algorithm \ref{alg:shuffling_process}). The teacher then samples tasks in the new task space using interpolation (see algorithm \ref{alg:interpolation_process}) which is now a discontinuous task space introducing peaks and cliffs in difficulty landscape. While the cut of regions is always the same, the shuffling process is seeded at each experiments.

\begin{algorithm}[H]
   \caption{Cutting and shuffling of the task space.}
   \label{alg:shuffling_process}
\begin{algorithmic}
   \STATE {\bfseries Input:} Number of dimensions $\mathcal{D}$, bounds $(min_i)_{i \in [\mathcal{D}]}$ and $(max_i)_{i \in [\mathcal{D}]}$, number of cuts $k$
   \FOR{$d \in [\mathcal{D}]$}
    \STATE Initialise arrays $\mathcal{O}_d, \mathcal{S}_d$
    \STATE $size \gets |max_d - min_d| / k$
        \FOR{$j \in [k]$}
            \STATE Store pair ($min_d + j*size$, $min_d + (j+1)*size$) in $\mathcal{O}_d$ and $\mathcal{S}_d$
        \ENDFOR
        \STATE Shuffle order of pairs in $\mathcal{S}_d$
    \ENDFOR
\end{algorithmic}
\end{algorithm}

\begin{algorithm}[H]
   \caption{Interpolate sampled task in the shuffled task space.}
    \label{alg:interpolation_process}
\begin{algorithmic}
   \STATE {\bfseries Input:} Number of dimensions $\mathcal{D}$, task vector $\mathcal{T}$, number of cuts $k$
   \STATE  Initialise the vector $\mathcal{I}$ of size $\mathcal{D}$
   \FOR{$d \in [\mathcal{D}]$}
        \FOR{$j \in [k]$}
            \STATE Get pair $o_j$ in $\mathcal{O}_d$
            \STATE Initialize $l$ with the first element of $o_j$
            \STATE Initialize $h$ with the second element of $o_j$
            \IF{$l \leq \mathcal{T}_d \leq h$}
                \STATE Get pair $s_j$ in $\mathcal{S}_d$
                \STATE Get $\beta$ as the interpolation of $\mathcal{T}_d$ from the interval $o_j$ to the interval $s_j$
                \STATE Set $\mathcal{I}_d = \beta$
                \STATE End the loop
            \ENDIF
        \ENDFOR
    \ENDFOR
    \RETURN $\mathcal{I}$
\end{algorithmic}
\end{algorithm}

\paragraph{Robustness to diverse students}
Finally, in order to highlight the robustness of an ACL teacher to diverse students, we perform $4$ experiments (each with $32$ seeds) and then aggregate results. We use the initial task space of Stump Tracks but use both PPO and SAC and two different embodiments:the short bipedal walker and the spider. Each embodiment is used both with PPO and SAC (hence $2$ experiments per embodiment and thus a total of $4$ experiments). We then aggregate the $128$ seeds into a single experiment result.

\subsection{Parkour experiments} \label{app:expe-details-parkour}
We perform a single experiment in the Parkour environment using $48$ seeds. Among these seeds, $16$ use a classic bipedal walker, $16$ a chimpanzee and $16$ a fish embodiment. We set the bounds of the task space to the following:
\begin{itemize}
    \item CPPN's input vector $\theta \in [-0.35, 0.05] \times [0.6, 1.0] \times [-0.1, 0.3]$
    \item Creepers' height $\mu_c \in [0; 4]$
    \item Creepers' spacing $\Delta_c \in [0; 5]$
    \item Water level $\tau \in [0; 1]$
\end{itemize}

Note that the above CPPN's input space is considered as our medium one. We also provide the easy space ($\theta \in [-0.25, -0.05] \times [0.8, 1.0] \times [0.0, 0.2]$) as well as the hard one ($\theta \in [-1, 1] \times [-1, 1] \times [-1, 1]$). Both the easy and medium spaces were designed from our hard task space. Their boundaries were searched such that the task space contains feasible tasks while maintaining diverse terrains. They differ in their ratio between feasible and unfeasible tasks. 

\subsubsection{Test sets}
\begin{figure}[H]
\vskip 0.2in
\begin{center}
\centerline{\includegraphics[width=\textwidth]{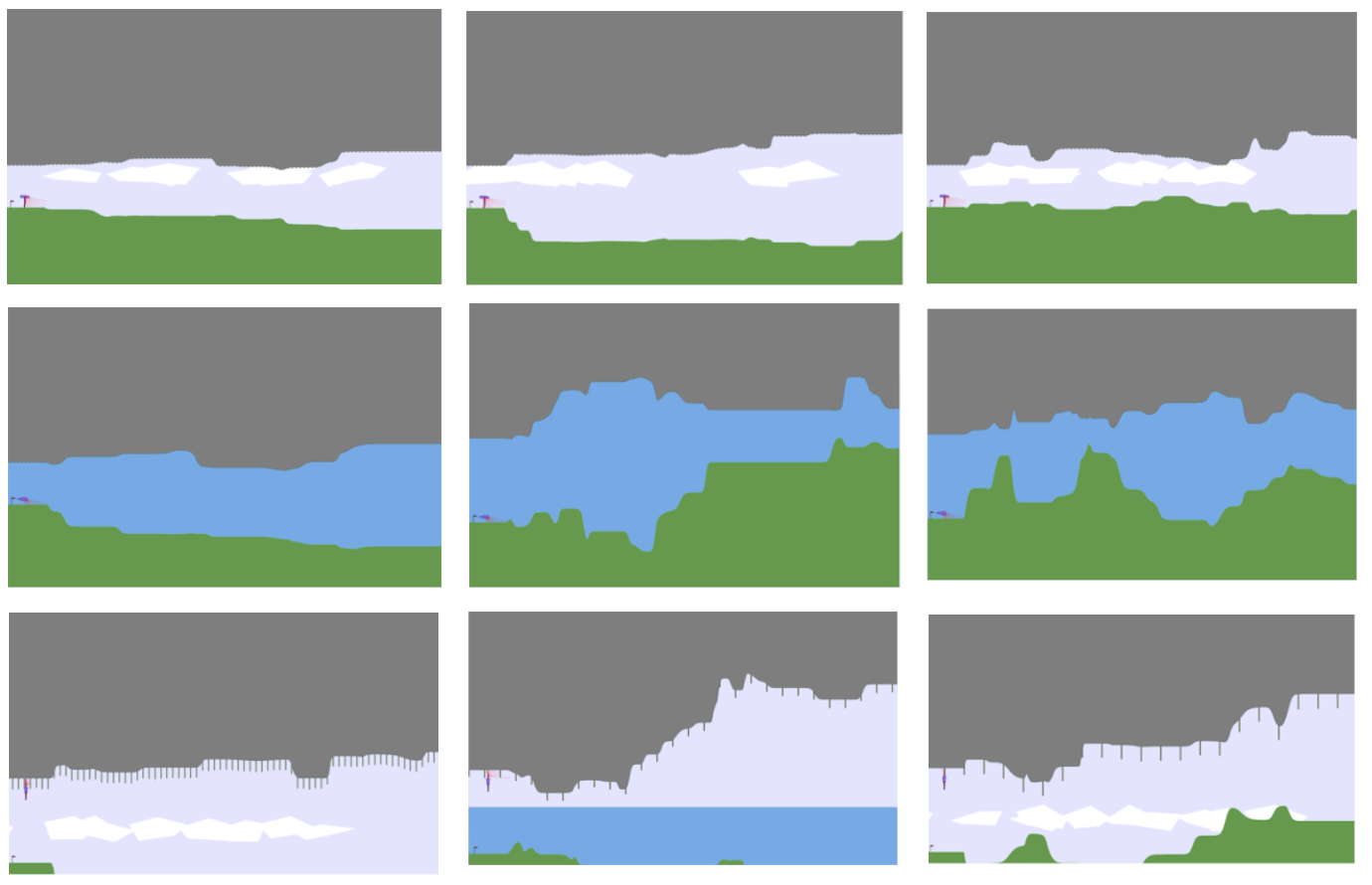}}
\caption{We show some examples of the tasks belonging to our Parkour's test sets. First line shows tasks from the walkers' test set, second line from the swimmers' one and finally last line for climbers. }
    \label{fig:Test_Sets}
\end{center}
\vskip -0.2in
\end{figure}

Unlike in the Stump Track experiments, we needed in the Parkour environment different test sets as our three embodiments (i.e.\ bipedal walker, chimpanzee, fish) are not meant to act and live in the same milieu (e.g.\ swimmers do not survive in tasks not containing water). Therefore, creating a test set composed of tasks uniformly sampled would not allow to assess the performance of the current embodiment. Hence, we made for the Parkour three different test sets, each constituted of $100$ tasks. As the task space previously defined is composed of mostly unfeasible tasks for any embodiment, we hand-designed each of the three test sets with the aim of showcasing the abilities of each morphology type, as well as showing the ability of the learned policy to generalize. Each test set has $60$ tasks that belong to the training task space and $40$ out-of-distribution tasks (using tasks outside the medium CPPN's input space as well as smoothing values different than $10$ for the $\delta$ parameter). They also share the same distribution between easy (1/3), medium (1/3) and hard (1/3) tasks. We chose each task such that it seems possible given the physical capacities of our embodiments. See figure \ref{fig:Test_Sets} for some examples of the test tasks.

%% file: appendices/additional_results.tex
\section{Additional results}\label{app:additional-results}
In this section, we provide additional results on experiments presented in section \ref{sec:experiments} as well as case studies. As mentioned in appendix \ref{app:expe-details}, we monitor both the episodic reward on each of the test tasks and the average episodic reward on training tasks every $500000$ steps for each seed. We use the episodic reward on test tasks to calculate our percentage of "mastered" tasks metric, which calculates the percentage of tasks on which the agent obtained an episodic reward greater than $230$. Additionally, we compare two algorithms in an experiment using Welch's t-test between their population of seeds. 

\subsection{Original Stump Tracks} \label{app:additional-results_original-stump-tracks}
We trained our SAC student for $20$ millions steps on the original Stump Track task space (i.e. $\mu_s \in [0; 3]$ and $\Delta_s \in [0;6]$) with each teacher and each expert knowledge setup. We used the best performance of each prior knowledge configuration as a baseline indication in figures \ref{fig:bench-vizu} and \ref{fig:profiling_5_millions}. As in our challenge-specific experiments, we used $32$ seeds as well as the same test set of $100$ evenly distributed tasks. Results can be found in table \ref{tab:baseline_results}.

\begin{table}[H]
\caption{Percentage of mastered tasks after $20$ millions steps on the original Stump Tracks challenge (i.e. $\mu_s \in [0;3]$ and $\Delta_s \in [0;6]$). Results shown are averages over $32$ seeds along with the standard deviation. We highlight the best results in bold, which then acted as an upper baseline indication in the challenge-specific comparisons.}
\label{tab:baseline_results}
\vskip 0.15in
\setlength\tabcolsep{4.5pt}
\begin{center}
\begin{small}
\begin{sc}
\begin{tabular}{lccc}
\toprule
\textbf{Algorithm} & No EK & Low EK & High EK\\ \midrule
ADR             & -                           & 24.1 ($\pm$ 20.8)            & 43.4 ($\pm$ 7.2)  \\
ALP-GMM         & \textbf{52.1} ($\pm$ 5.9)   & \textbf{47.1} ($\pm$ 13.9)   & 49.3 ($\pm$ 5.9)  \\
Covar-GMM       & 43.0 ($\pm$ 9.1)            & 40.25 ($\pm$ 16.5)           & 45.2($\pm$ 10.1)  \\
GoalGAN         & -                           & 29.9 ($\pm$ 26.2)            & \textbf{51.9} ($\pm$ 7.3)  \\
RIAC            & 40.5 ($\pm$ 8.4)            & 39.6 ( ($\pm$ 11.2)          & 42.2 ($\pm$ 5.4)  \\
SPDL            & 20.8 ($\pm$ 19.4)           & 18.5 ($\pm$ 20.8)            & 34.0 ($\pm$ 10.6)  \\
Setter-Solver   & 25.3 ($\pm$ 10.7)           & 36.6($\pm$ 10.2)             & 37.4 ($\pm$ 9.8)  \\ \bottomrule
 \end{tabular}
\end{sc}
\end{small}
\end{center}
\vskip -0.1in
\end{table}

\subsection{Challenge-specific comparison} \label{app:additional-results_stump-tracks}
\subsubsection{Overall results}
\begin{figure}[H]
\vskip 0.2in
\begin{center}
\centerline{\includegraphics[width=0.9\textwidth]{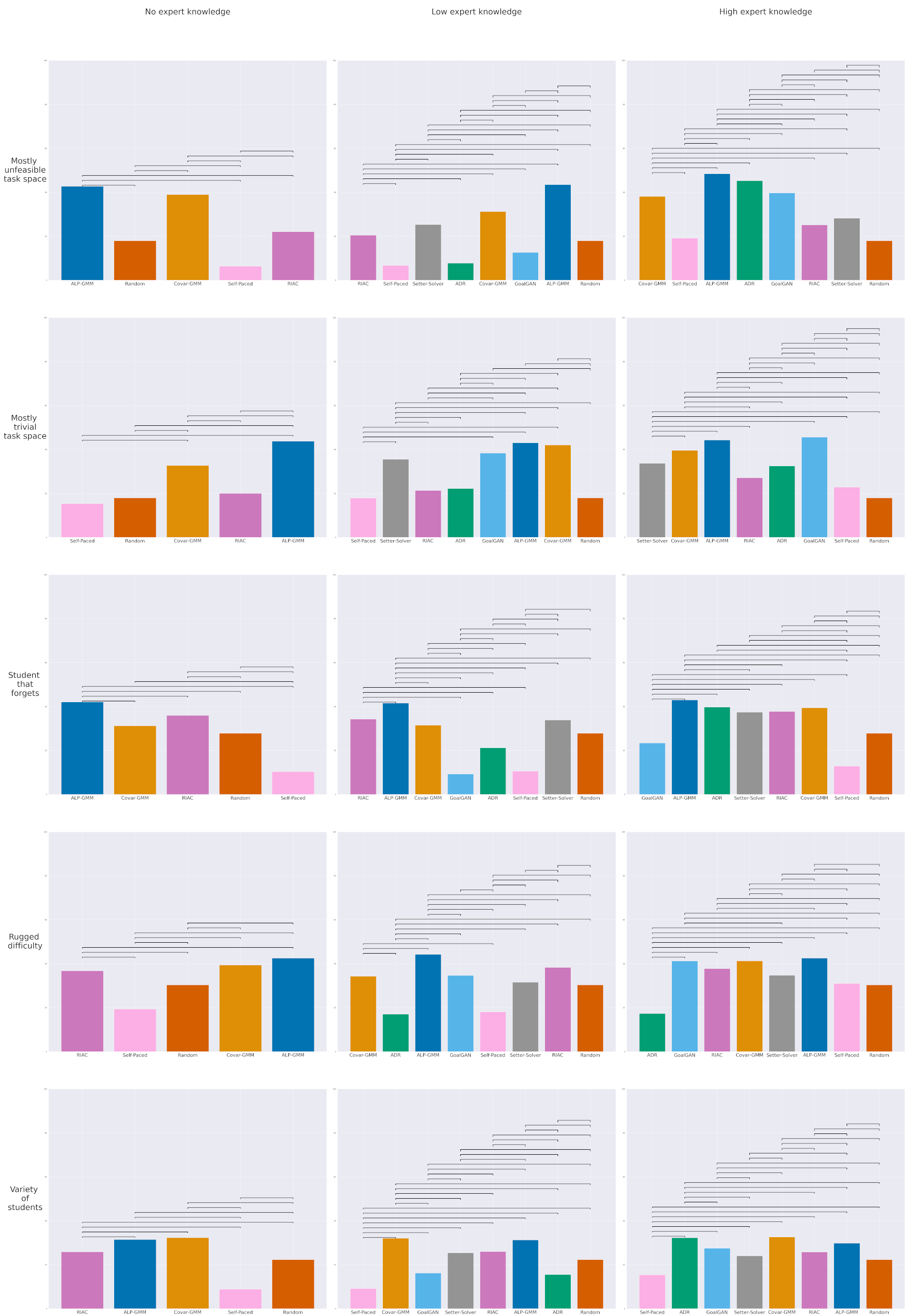}}
\caption{Performance of the different teachers at the end of training in every experiment of our challenge-specific comparison. We plot as bars the average percentage of mastered tasks for each ACL method. Additionally, we compare in every experiment all possible couples of teacher methods using Welch's t-test and annotate the significantly different ($p<0.05$) ones.}
 \label{fig:profiling_bars}
\end{center}
\vskip -0.2in
\end{figure}

We here show the performance after $20$ millions steps of each ACL teacher on each challenge. Results are gathered in tables \ref{tab:no_ek_results}, \ref{tab:low_ek_results} and \ref{tab:high_ek_results}, as well as in figure \ref{fig:profiling_bars} where we show the results of Welch's t-test between all methods on every challenge.
\clearpage

\begin{table}[H]
\caption{Percentage of mastered tasks after $20$ millions steps with \textbf{no} prior knowledge in each challenge. Results shown are averages over all seeds along with the standard deviation. We highlight the best results in bold.}
  \label{tab:no_ek_results}
\vskip 0.15in
\setlength\tabcolsep{4.5pt}
\begin{center}
\begin{small}
\begin{sc}
\begin{tabular}{lccccc}
\toprule
\textbf{Algorithm} & Mostly unf. & Mostly triv. & Forgetting stud. & Rugged dif. & Diverse stud.          \\ \midrule
Random          & 18.0 ($\pm$ 10.5)   & 22.2 ($\pm$ 15.2)    & 27.8 ($\pm$ 14.6)  & 30.3 ($\pm$ 7.7)    & 22.3 ($\pm$ 11.5)  \\
ALP-GMM         & \textbf{42.8} ($\pm$ 6.6)   & \textbf{43.7} ($\pm$ 6.0)    & \textbf{42.1} ($\pm$ 6.9)  & \textbf{42.5} ($\pm$ 4.8)    & 31.5 ($\pm$ 9.2)  \\
Covar-GMM       & 39.0 ($\pm$ 9.9)   & 32.7 ($\pm$ 16.0)    & 31.3 ($\pm$ 16.2)  & 39.4 ($\pm$ 7.4)    & \textbf{32.3} ($\pm$ 10.6)  \\
RIAC            & 22.1 ($\pm$ 14.5)    & 20.0 ($\pm$ 10.9)   & 36.8 ($\pm$ 6.9)  & 36.4 ($\pm$ 7.9)    & 25.9 ($\pm$ 11.3)  \\
SPDL            & 6.4 ($\pm$ 10.2)   & 15.3 ($\pm$ 9.9)    & 10.4 ($\pm$ 12.9)  & 19.3 ($\pm$ 16.2)    & 8.9 ($\pm$ 14.4)  \\ \bottomrule
 \end{tabular}
\end{sc}
\end{small}
\end{center}
\vskip -0.1in
\end{table}

\begin{table}[H]
\caption{Percentage of mastered tasks after $20$ millions steps with \textbf{low} prior knowledge in each challenge. Results shown are averages over all seeds along with the standard deviation. We highlight the best results in bold.}
\label{tab:low_ek_results}
\vskip 0.15in
\setlength\tabcolsep{4.5pt}
\begin{center}
\begin{small}
\begin{sc}
\begin{tabular}{lccccc}
\toprule
\textbf{Algorithm} & Mostly unf. & Mostly triv. & Forgetting stud. & Rugged dif. & Diverse stud.          \\ \midrule
Random          & 18.0 ($\pm$ 10.1)   & 18.0 ($\pm$ 7.1)    & 27.8 ($\pm$ 14.6)  & 30.3 ($\pm$ 7.7)    & 22.3 ($\pm$ 11.5)  \\
ADR             & 7.8 ($\pm$ 17.9)   & 22.2 ($\pm$ 15.2)    & 21.2 ($\pm$ 21.2)  & 17.0 ($\pm$ 19.6)    & 15.6 ($\pm$ 19.1)  \\
ALP-GMM         & \textbf{43.5} ($\pm$ 13.0)   & \textbf{43.0} ($\pm$ 9.0)    & \textbf{41.6} ($\pm$ 12.5)  & \textbf{44.2} ($\pm$ 7.1)   & 31.3 ($\pm$ 9.4)  \\
Covar-GMM       & 31.2 ($\pm$ 16.8)   & 42.0 ($\pm$ 8.4)    & 31.5 ($\pm$ 18.4)  & 34.3 ($\pm$ 10.7)    & \textbf{32.1} ($\pm$ 9.6)  \\
GoalGAN         & 12.7 ($\pm$ 16.2)   & 38.4 ($\pm$ 16.1)    & 9.3 ($\pm$ 15.8)  & 34.7 ($\pm$ 19.1)    & 16.2 ($\pm$ 17.5)  \\
RIAC            & 20.5 ($\pm$ 14.0)   & 21.3 ($\pm$ 8.8)    & 34.3 ($\pm$ 12.5)  & 38.3 ($\pm$ 11.3)    & 26.0 ($\pm$ 11.7)  \\
SPDL            & 6.7 ($\pm$ 10.2)   & 17.9 ($\pm$ 12.2)    & 10.6 ($\pm$ 12.2)  & 18.1 ($\pm$ 15.8)    & 9.2 ($\pm$ 14.2)  \\
Setter-Solver   & 25.3 ($\pm$ 10.7)   & 35.5 ($\pm$ 8.9)    & 33.9 ($\pm$ 12.5)  & 31.6 ($\pm$ 11.3)    & 25.4 ($\pm$ 9.0)  \\ \bottomrule
 \end{tabular}
\end{sc}
\end{small}
\end{center}
\vskip -0.1in
\end{table}

\begin{table}[H]
\caption{Percentage of mastered tasks after $20$ millions steps with \textbf{high} prior knowledge in each challenge. Results shown are averages over all seeds along with the standard deviation. We highlight the best results in bold.}
\label{tab:high_ek_results}
\vskip 0.15in
\setlength\tabcolsep{4.5pt}
\begin{center}
\begin{small}
\begin{sc}
\begin{tabular}{lccccc}
\toprule
\textbf{Algorithm} & Mostly unf. & Mostly triv. & Forgetting stud. & Rugged dif. & Diverse stud.          \\ \midrule
Random          & 18.0 ($\pm$ 10.1)   & 18.0 ($\pm$ 7.1)    & 27.8 ($\pm$ 14.6)  & 30.3 ($\pm$ 7.7)    & 22.3 ($\pm$ 11.5)  \\
ADR             & 45.3 ($\pm$ 6.7)   & 32.5 ($\pm$ 6.2)    & 39.8 ($\pm$ 10.8)  & 17 ($\pm$ 20.9)    & 32.3 ($\pm$ 9.7)  \\
ALP-GMM         & \textbf{48.4} ($\pm$ 11.2)   & 44.3 ($\pm$ 14.2)    & \textbf{43.0} ($\pm$ 9.0)  & \textbf{42.5} ($\pm$ 7.3)    & 29.8 ($\pm$ 8.8)  \\
Covar-GMM       & 38.2 ($\pm$ 11.9)   & 39.6 ($\pm$ 10.3)    & 39.5 ($\pm$ 12.5)  & 41.3 ($\pm$ 7.0)    & \textbf{32.6} ($\pm$ 10.2)  \\
GoalGAN         & 39.7 ($\pm$ 10.1)   & \textbf{45.6} ($\pm$ 13.5)    & 23.4 ($\pm$ 19.7)  & 41.2 ($\pm$ 12.6)    & 27.5 ($\pm$ 9.4)  \\
RIAC            & 25.2 ($\pm$ 12.3)   & 22.1 ($\pm$ 11.1)    & 37.7 ($\pm$ 12.5)  & 37.7 ($\pm$ 8.8)    & 25.8 ($\pm$ 11.7)  \\
SPDL            & 19.1 ($\pm$ 12.5)   & 22.9 ($\pm$ 6.9)    & 12.9 ($\pm$ 11.2)  & 31.0 ($\pm$ 11.2)    & 15.4 ($\pm$ 15.1)  \\
Setter-Solver   & 28.2 ($\pm$ 9.7)   & 33.7 ($\pm$ 10.8)    & 37.4 ($\pm$ 8.7)  & 34.7 ($\pm$ 8.1)    & 24.0 ($\pm$ 9.8)  \\ \bottomrule
 \end{tabular}
\end{sc}
\end{small}
\end{center}
\vskip -0.1in
\end{table}
 
\subsubsection{Case study: Sample efficiency}
\begin{figure}[H]
\vskip 0.2in
\begin{center}
\centerline{\includegraphics[width=\textwidth]{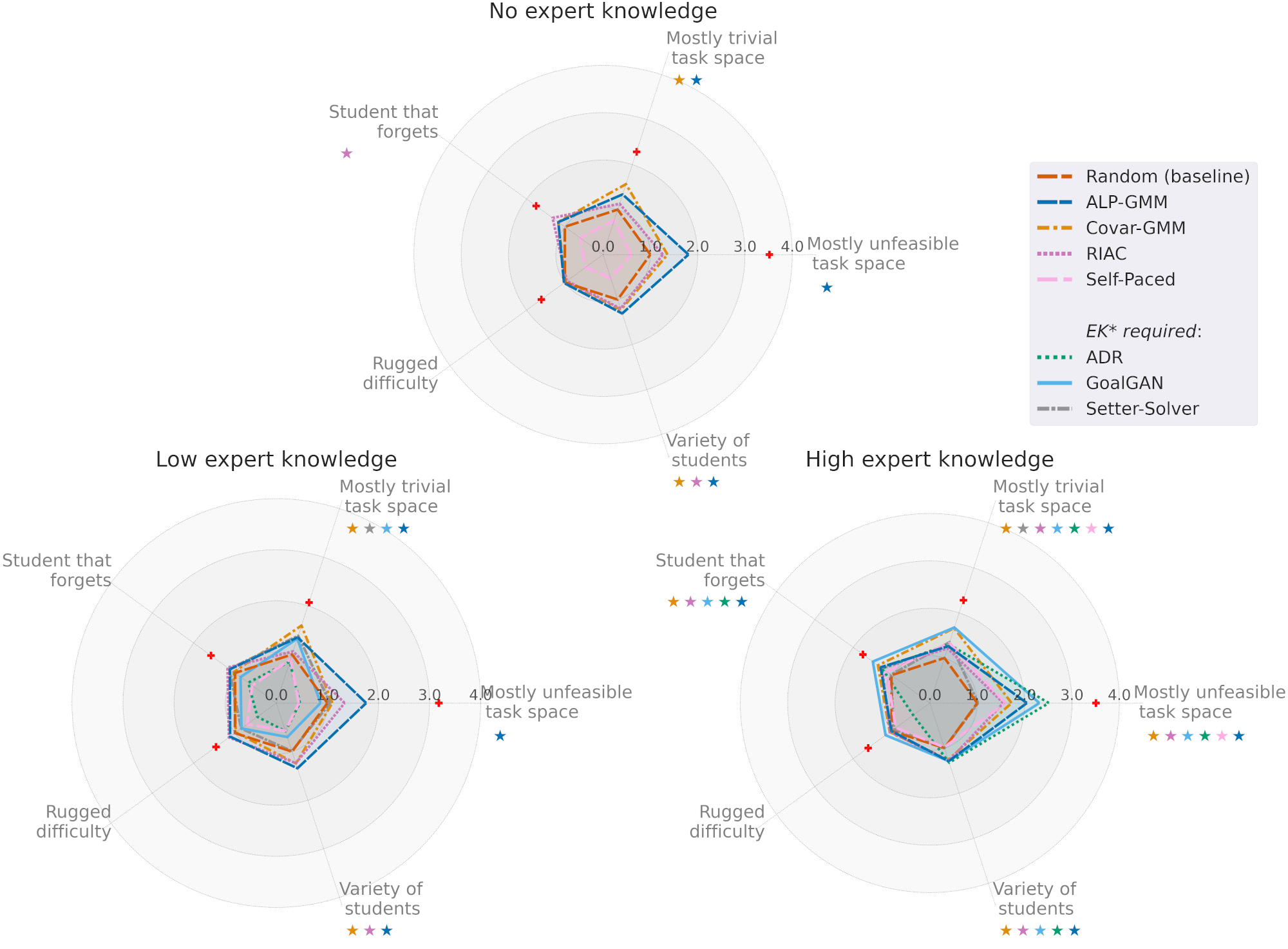}}
\caption{\textbf{Performance of ACL methods measured after $5$ millions steps.} 
Results are presented as an order of magnitude of the performance of Random. Performance is defined as the average percentage of mastered test tasks over all $32$ seeds. We also provide the same indications ({\textcolor{red}{\ding{58}}}) of the best performance (measured after $5$ millions steps) on the original Stump Tracks experiment as in figure \ref{fig:bench-vizu}. Finally, we indicate on each axis which method performed significantly better than Random ($p<0.05$) using colored stars matching each method's color (e.g. {\textcolor{Covar_orange}{\ding{72}}} for Covar-GMM, {\textcolor{ADR_green}{\ding{72}}} for ADR). \textit{EK: Expert Knowledge.}}
\label{fig:profiling_5_millions}
\end{center}
\vskip -0.2in
\end{figure}

In this section, we take a look at the sample efficiency of the different ACL methods using their performance after only $5$ millions steps. We reuse the same radar chart as in section \ref{sec:experiments} in figure \ref{fig:profiling_5_millions}. 

Looking at results, one can see the impact of ACL in the mostly unfeasible challenge, as some methods (e.g. ALP-GMM or ADR with high expert knowledge) already reach twice the performance of random after only $5$ million steps. This highlights how leveraging a curriculum adapted to the student's capabilities is key when most tasks are unfeasible. On the opposite, when the task space is easier (as in the mostly trivial challenge), Random samples more tasks suited for the current student's abilities and the impact of Curriculum Learning is diminished. 

Having the difficulty landscape rugged makes the search for learnable and adapted subspaces harder. Figure \ref{fig:profiling_5_millions} shows that only $5$ millions steps is not enough, even for teachers like ALP-GMM or Covar-GMM theoretically more suited for rugged difficulty landscapes, to explore and leverage regions with high learning progress. 

Finally, one can see the strong impact of a well set initial distribution of tasks in the beginning of learning. Indeed, both ADR and GoalGAN already almost reach their final performance (i.e. the one they reached after $20$ millions steps shown in figure \ref{fig:bench-vizu}) after $5$ millions steps in the \textit{High expert knowledge} setup, as they know where to focus and do not need exploration to find feasible subspaces. Similarly, adding expert knowledge to ALP-GMM increases its performance compared to the no and low expert knowledge setups, helping it focus the bootstrapping process on a feasible region. Leveraging this initial task distribution, GoalGAN obtains the best results in 3/5 challenges after $5$ millions steps with high expert knowledge. This shows, in addition of the results from section \ref{sec:experiments}, that GoalGAN is a very competitive method, especially when it has access to high expert knowledge.
 
 \subsubsection{Case study: On the difficulty of GoalGAN and SPDL to adapt the curriculum to forgetting students}
\begin{figure}[H]
\vskip 0.2in
\begin{center}
\centerline{\includegraphics[width=\textwidth]{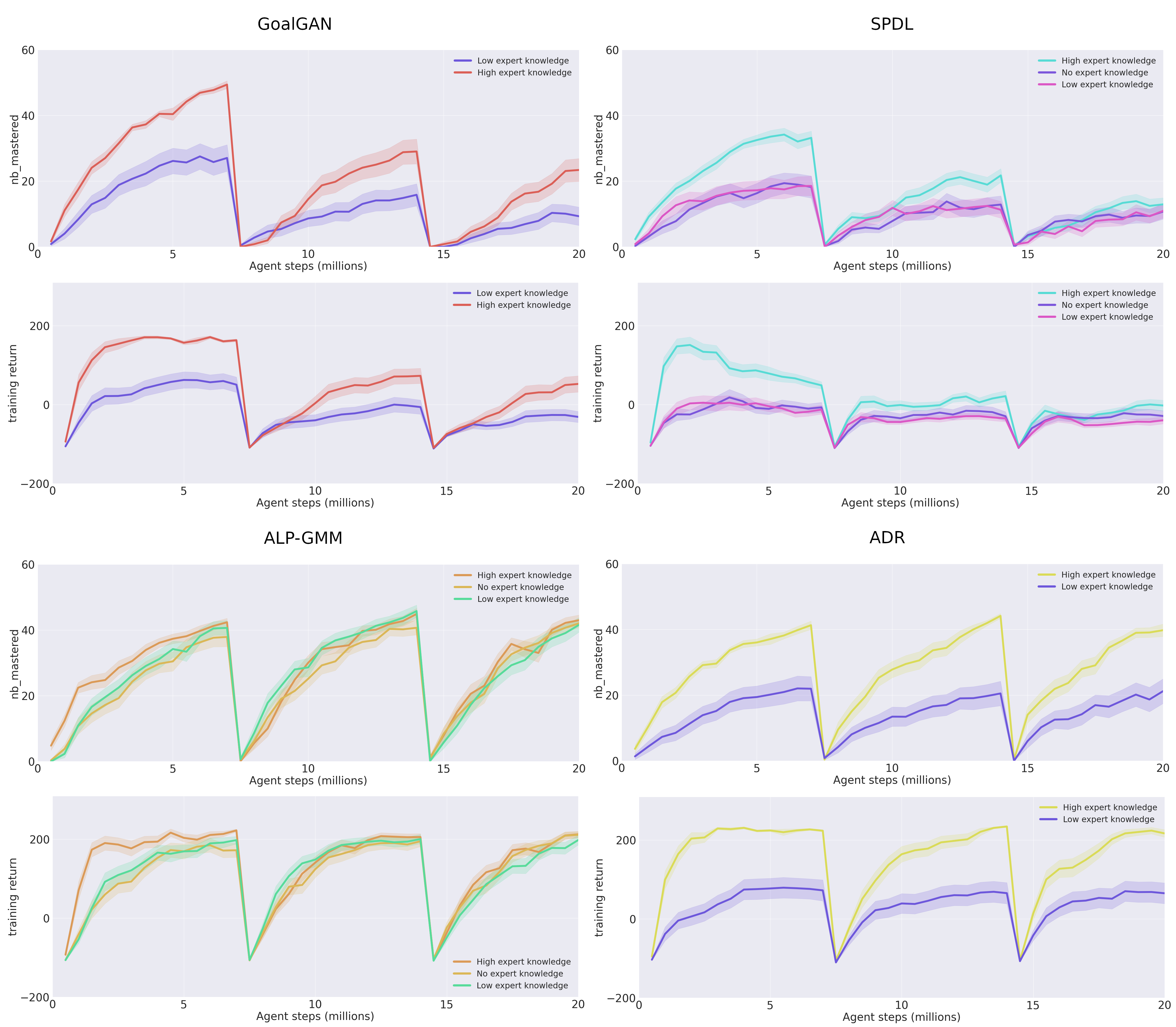}}
\caption{Percentage of mastered test tasks and average training return of GoalGAN, SPDL, ALP-GMM and ADR on the forgetting student challenge. Curves are averages over $32$ seeds along with the standard error of the mean.}
\label{fig:criteria_3}
\end{center}
\vskip -0.2in
\end{figure}

 As mentioned in section \ref{sec:experiments}, both GoalGAN and SPDL struggled on the forgetting student challenge, no matter the amount of expert knowledge. In order to better understand their behaviour in this challenge, we plot in figure \ref{fig:criteria_3} both the evolution of their percentage of mastered tasks and their average training return. We also add ALP-GMM and ADR (two students that performed well in this challenge) as baselines for comparison. While ADR and ALP-GMM make the student quickly recover from a reset (i.e. the percentage of mastered tasks quickly reaches the performance it had before the reset) and then carry on improving, both GoalGAN and SPDL suffer from resets and do not manage to recover, leading to a poor final performance.
 
This phenomenon could be explained by multiple factors. First, in the case of SPDL, even though the algorithm tries to shift its sampling distribution such that it maximizes the student's performance, the optimization methods also has to minimize the distance to the target distribution, which is a Gaussian spanning over the entire task space. However, resetting the student's policy requires the ACL method to revert back to the initial simple task distribution that it proposed at the beginning of training. Such a reverse process might not easily be achievable by SPDL, which optimization procedure progressively shifts its sampling distribution towards the target one.

Concerning GoalGAN, the performance impact of student resets is most likely due to its use of a buffer of "Goals of Intermediate Difficulty", used to train the goal generator. Upon student reset, this buffer becomes partially obsolete, as the student is reinitialized, making it lose all learned walking gaits, i.e. everything must be learned again. This means the goal generator will propose tasks that are too complex for a student that is just starting to learn. Because GoalGAN cannot reset its buffer of "Goals of Intermediate Difficulty" (which would require knowledge over the student's internal state), it impairs its ability to quickly shift to simpler task subspaces.

 \subsubsection{Case study: Impact of expert knowledge on ALP-GMM}
\begin{figure}[H]
\vskip 0.2in
\begin{center}
\centerline{\includegraphics[width=\textwidth]{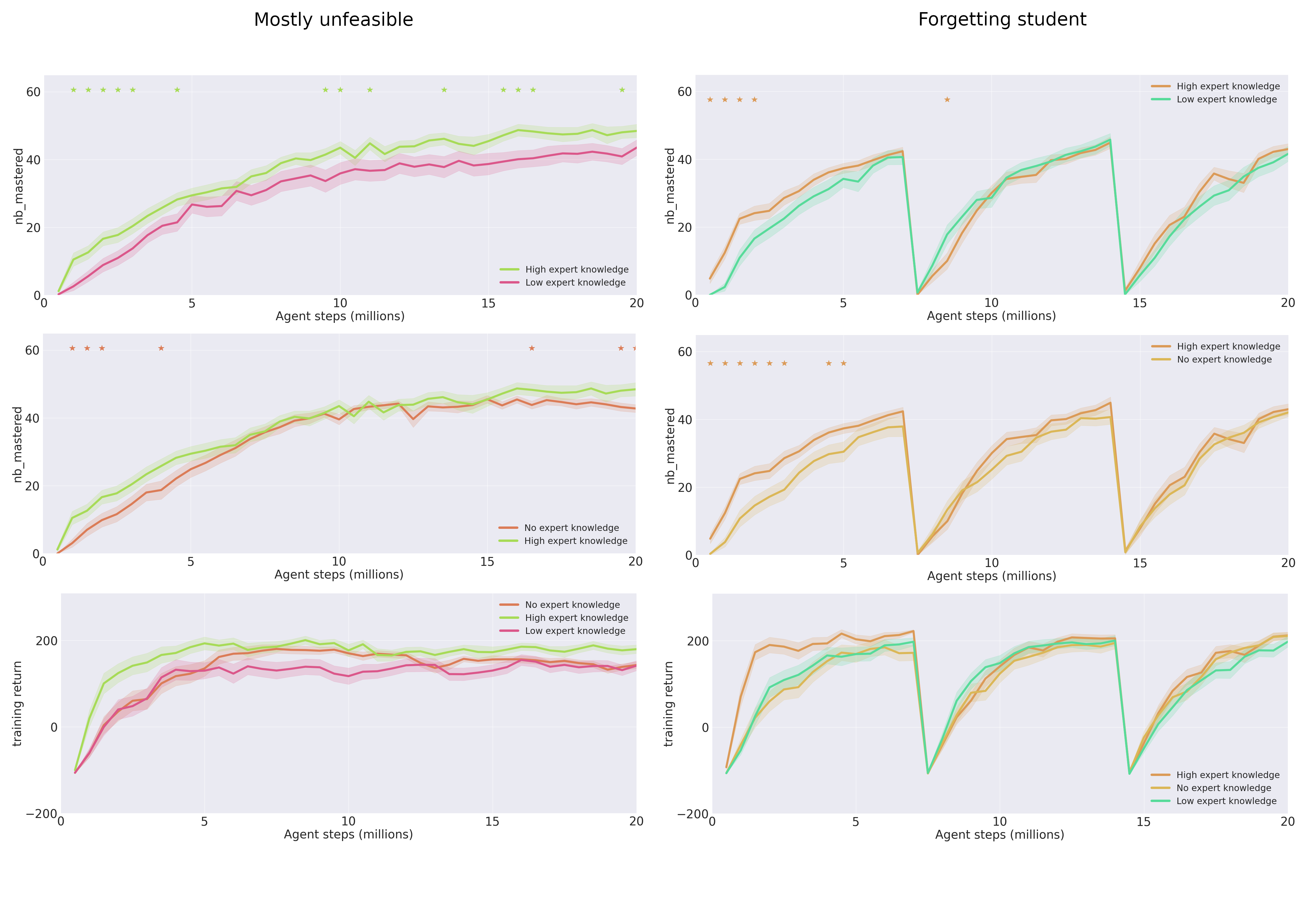}}
\caption{Percentage of mastered test tasks and average training return of ALP-GMM on both the mostly unfeasible and the forgetting student challenges. Curves are averages over $32$ seeds along with the standard error of the mean. We compare the impact of high expert knowledge compared to the two other setups using Welch's t-test and highlight significant ($p<0.05$) differences with stars.)}
\label{fig:alp_ek}
\end{center}
\vskip -0.2in
\end{figure}

 As aforementioned, ALP-GMM is a method initially not requiring any expert knowledge. Moreover, it relies on an exploration (bootstrap) phase to fill its buffer, usually using uniform sampling over the task space. In \benchname, we provide an extended version of it where we added the possibility to use expert knowledge by bootstrapping from an initial distribution instead of a uniform distribution. In this case study, we take a look at the impact such a theoretical improvement had on their performance. We focus on the mostly unfeasible and forgetting student challenges, as the first highlighted the most how prior knowledge can help an ACL method (helping it start in a feasible region) and the latter showed interesting results for this case study, in addition of being easy to analyse (as it only uses a bipedal walker on the original task space of the Stump Tracks). We gather these results in figure \ref{fig:alp_ek}. Note that both the no and low expert knowledge setups are the same for ALP-GMM , meaning that any difference between their results is only due to variance in both the student's learning and ACL process.
 
 When looking at these results, one can see that the high expert knowledge setup is significantly better than the two other setups at the beginning of the training in both challenges. These results can also be completed by our sample efficiency case study (see figure \ref{fig:profiling_5_millions}), showing that adding expert knowledge to ALP-GMM makes it more sample efficient. Then, as training advances, the difference becomes statistically insignificant ($p>0.05$). Finally, while the final results given in tables \ref{tab:no_ek_results}, \ref{tab:low_ek_results}, and \ref{tab:high_ek_results} show an improved percentage of mastered tasks in almost all challenges, with a notable difference (at least $+5$) on the mostly unfeasible challenge, results on the original Stump Tracks experiments (table \ref{tab:baseline_results}) show better results with no expert knowledge. It is thus not clear whether adding this prior knowledge to ALP-GMM benefits the whole training instead of just the beginning. Note that similar behaviours were also obtained with Covar-GMM, even though they were not as significant as the ones of ALP-GMM.
 
 \subsubsection{Case study: What ADR needs}
\begin{figure}[H]
\vskip 0.2in
\begin{center}
\centerline{\includegraphics[width=\textwidth]{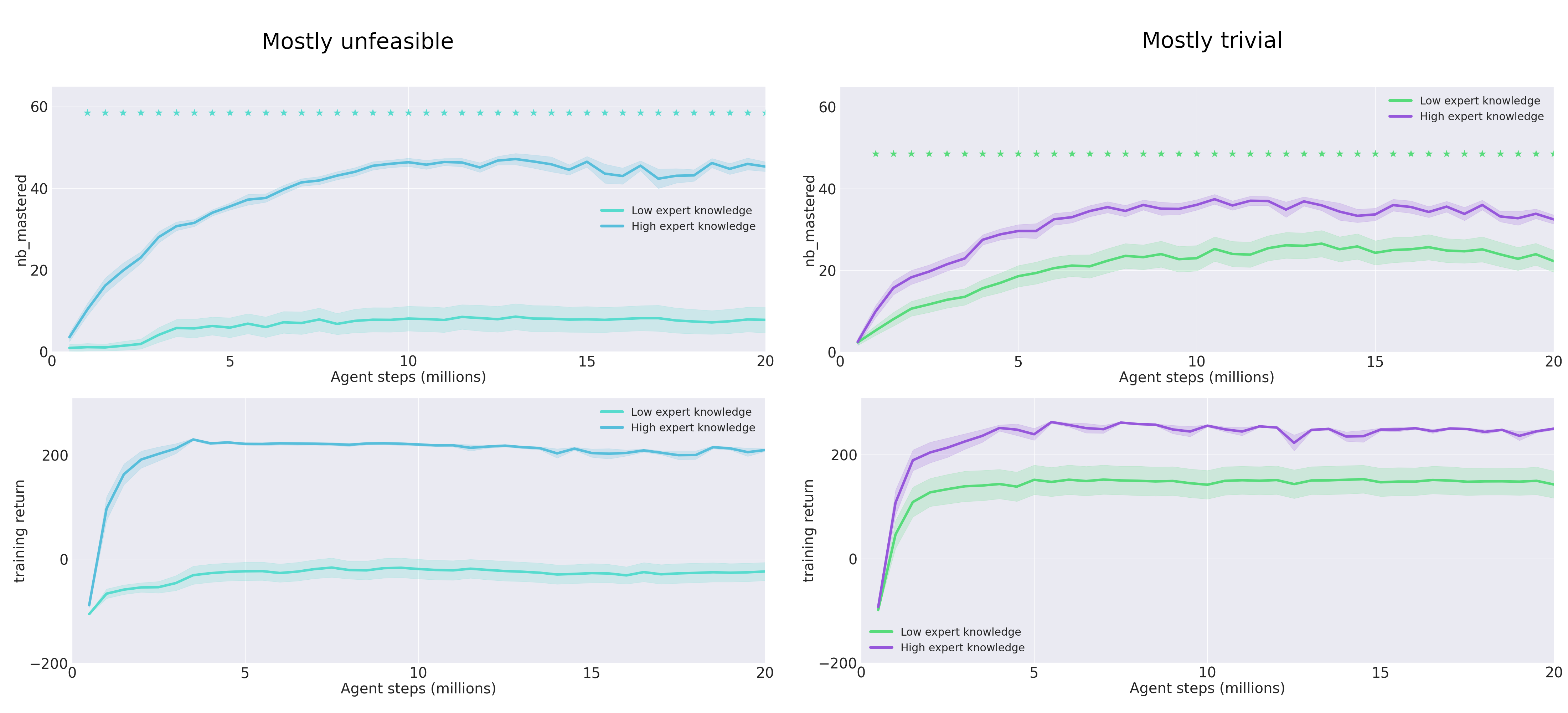}}
\caption{Percentage of mastered test tasks and average training return of ADR on both the mostly unfeasible and mostly trivial challenges. Curves are averages over $32$ seeds along with the standard error of the mean. We compare the impact of high to low expert knowledge using Welch's t-test ($p<0.05$) and highlight significant differences with stars.)}
\label{fig:adr_ek}
\end{center}
\vskip -0.2in
\end{figure}

 ADR is a very efficient and light method, that, when all its expert knowledge requirements are satisfied, competes with the best teachers. However, in order to obtain such an efficient behaviour, ADR needs certain conditions that are implied by its construction. First, as explained in appendix \ref{app:acl-details}, ADR starts its process using a single task, and makes the assumption that this latter is easy enough for the freshly initialized student. It then progressively grows its sampling distribution around this task if the student manages to "master" the proposed tasks. While this behaviour seems close to SPDL's, ADR does not have any target distribution to help it shift the distribution even if the student's performance are not good enough. Hence, ADR can get completely stuck if it is initialized on a task lying in a very hard region, whereas SPDL would still try to converge to the target distribution (even though the performance would not be as good as if its initial distribution was set in an easy subspace). Similarly, GoalGAN also uses an initial distribution at the beginning of the training which, as shown in the results, has a strong impact on the final performance. However, even without it, GoalGAN is still able to reach a decent performance in certain challenges (e.g. mostly trivial) unlike ADR. This observation can also be seen in the Parkour's experiments, where GoalGAN reaches the top $4$ while ADR obtains the worst performance. In order to highlight this explanation, we provide the results of ADR in the mostly unfeasible and mostly trivial challenges in figure \ref{fig:adr_ek}, in addition of the clear difference between expert knowledge setups showed by figure \ref{fig:bench-vizu} and tables \ref{tab:no_ek_results}, \ref{tab:low_ek_results}, and \ref{tab:high_ek_results}. Using figure \ref{fig:adr_ek}, one can see the clear and significant ($p<0.05$) difference between the two expert knowledge setups.
 
\begin{figure}[H]
\vskip 0.2in
\begin{center}
\centerline{\includegraphics[width=\textwidth]{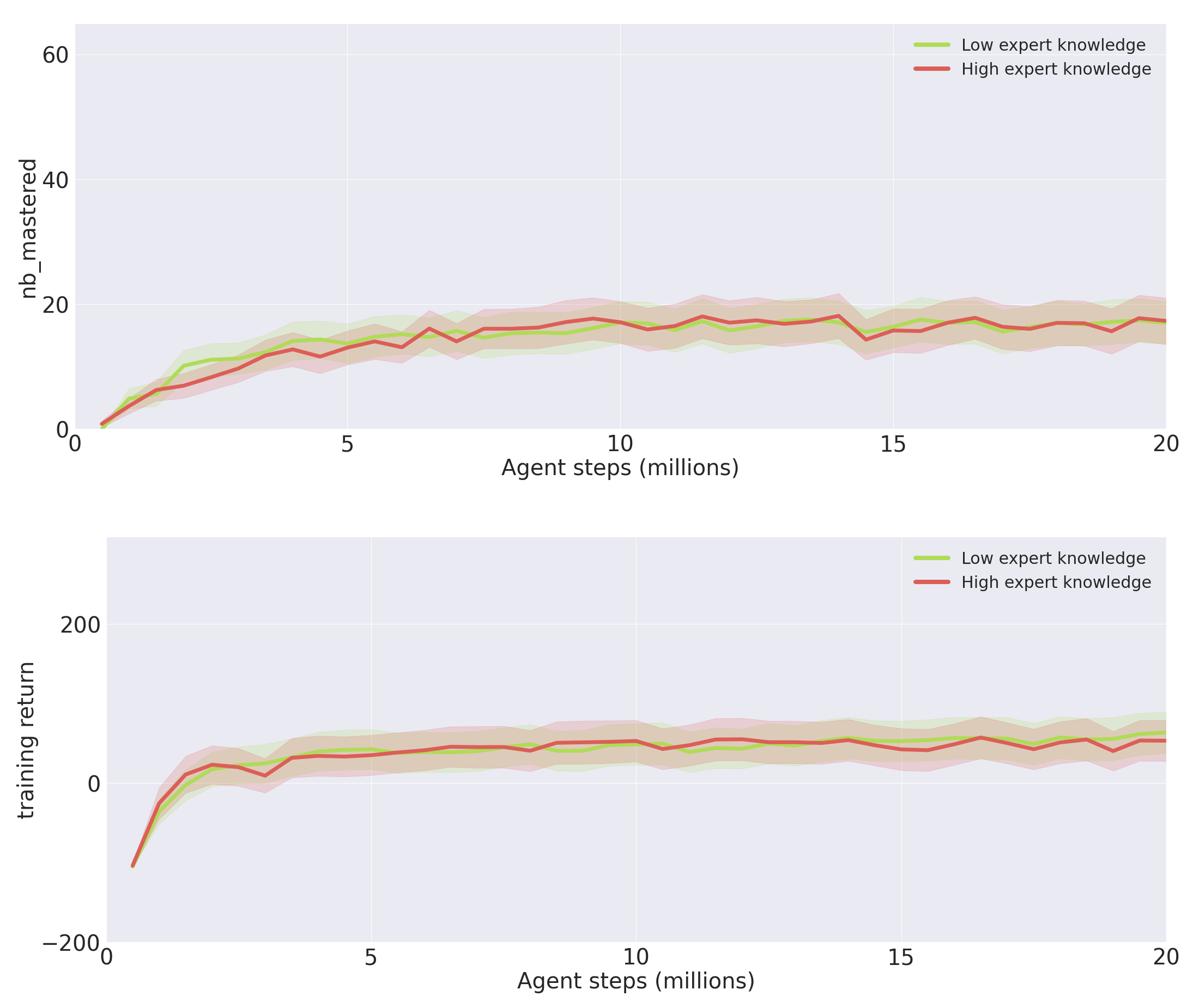}}
 \caption{Percentage of mastered test tasks and average training return of ADR on rugged difficulty landscape challenge. Curves are averages over $32$ seeds along with the standard error of the mean.}
 \label{fig:adr_rugged_dif}
\end{center}
\vskip -0.2in
\end{figure}

 In addition of an initial task well set using prior knowledge about the task space, ADR needs a difficulty landscape not too rugged to be able to expand and reach regions with high learning progress for the student. Indeed, when looking at its algorithm, one can see that the sampling distribution grows in a certain direction only if the student is able to master the tasks proposed at the edge of the distribution on this direction. If it is not the case (i.e. if this region of the task space is too hard for the current student's capabilities), the sampling distribution will shrink. This simple mechanism makes the strong assumption that if the difficulty is too hard at one edge of the distribution, there is no need to go further, implicitly saying that the difficulty further is at least as hard as the one at the edge. While this works well in the vanilla task space of our Stump Tracks environment (our difficulty is clearly smooth and even monotonically increasing), any task space with a rugged difficulty landscape would make the problem harder for ADR. Indeed, as it reaches a valley in the difficulty landscape surrounded by hills of unfeasible (or too hard for the current student's abilities) tasks, ADR can get stuck. In order to highlight this behaviour, we use our rugged difficulty landscape challenge, where we created a discontinuous difficulty landscape where unfeasible regions can lie in the middle of the task space. Figure \ref{fig:adr_rugged_dif} shows how ADR is unable to solve this challenge, no matter the amount of expert knowledge it uses, leading to the worst performance of our benchmark (significantly worse than Random at $p<0.05$). Note that this issue also happens in our Parkour experiments, as the difficulty of the task space is very rugged (see section \ref{sec:experiments}).

\subsection{Parkour}\label{app:additional-results_parkour}
 \subsubsection{Overall results}
\begin{figure}[H]
\vskip 0.2in
\begin{center}
\centerline{\includegraphics[width=0.95\textwidth]{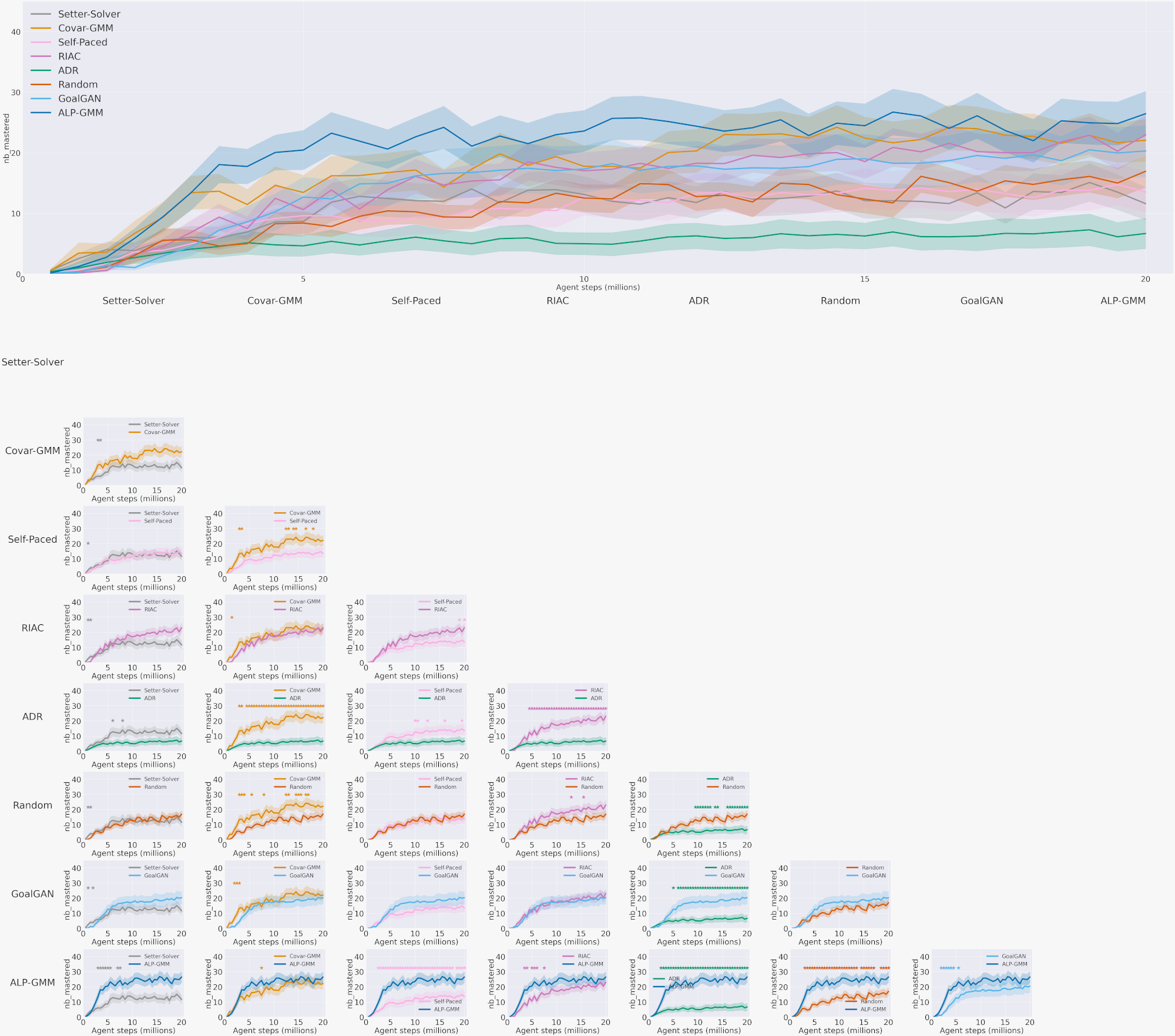}}
\caption{\textbf{Comparison of the ACL methods on the Parkour experiments}. Upper figure shows the average percentage of mastered tasks over the $48$ seeds with the standard error of the mean. We then extract all the possible couples of methods and compare their two curves. At each time step (i.e. every $500000$ steps), we use Welch's t-test to compare the two distributions of seeds. If a significant difference exists ($p<0.05$), we add a star above the curves at this time step.}
\label{fig:full_parkour_comparison}
\end{center}
\vskip -0.2in
\end{figure}

 In this section, we present the performance of our teacher algorithms on the Parkour track experiments.
 We present the final results in table \ref{tab:parkour_results} as well as a comparison in figure  \ref{fig:full_parkour_comparison} using Welch's t-test. We also provide insights concerning the obtained policies at  \url{http://developmentalsystems.org/TeachMyAgent/}. When looking at the overall results, one can see that ALP-GMM is the only teacher performing significantly better than Random throughout training. Covar-GMM's performance are very close to ALP-GMM, as well as RIAC, which obtains very similar results to GoalGAN. While being very close to Random, Setter-Solver's results are not significantly different from ALP-GMM's results by the end of the training. Finally, while SPDL's behavior is very similar to Random, ADR reaches a plateau very soon and eventually obtains significantly worse results than the random teacher. 
 
\begin{table}[H]
\caption{Percentage of mastered tasks after $20$ millions steps on the Parkour track. Results shown are averages over $16$ seeds along with the standard deviation for each morphology as well as the aggregation of the $48$ seeds in the overall column. We highlight the best results in bold.}
\label{tab:parkour_results}
\vskip 0.15in
\setlength\tabcolsep{4.5pt}
\begin{center}
\begin{small}
\begin{sc}
\begin{tabular}{lcccc}
\toprule
\textbf{Algorithm} & BipedalWalker & Fish & Climber & Overall \\ \midrule
Random          & 27.25 ($\pm$ 10.7)            & 23.6 ($\pm$ 21.3)             & 0.0 ($\pm$ 0.0)               & 16.9 ($\pm$ 18.3)  \\
ADR             & 14.7 ($\pm$ 19.4)             & 5.3 ($\pm$ 20.6)              & 0.0 ($\pm$ 0.0)               & 6.7 ($\pm$ 17.4)  \\
ALP-GMM         & \textbf{42.7} ($\pm$ 11.2)    & 36.1 ($\pm$ 28.5)             & 0.4 ($\pm$ 1.2)               & \textbf{26.4} ($\pm$ 25.7)  \\
Covar-GMM       & 35.7 ($\pm$ 15.9)             & 29.9 ($\pm$ 27.9)             &  0.5 ($\pm$ 1.9)              & 22.1 ($\pm$ 24.2)  \\
GoalGAN         & 25.4 ($\pm$ 24.7)             & 34.7 ($\pm$ 37.0)             & 0.8 ($\pm$ 2.7)               & 20.3 ($\pm$ 29.5)  \\
RIAC            & 31.2 ($\pm$ 8.2)              & \textbf{37.4} ($\pm$ 25.4)    & 0.4  ($\pm$ 1.4)              & 23.0 ($\pm$ 22.4)  \\
SPDL            & 30.6 ($\pm$ 22.8)             & 9.0 ($\pm$ 24.2)              & \textbf{1.0} ($\pm$ 3.4)      & 13.5 ($\pm$ 23.0)  \\
Setter-Solver   & 28.75 ($\pm$ 20.7)            & 5.1 ($\pm$ 7.6)               & 0.0 ($\pm$ 0.0)               & 11.3 ($\pm$ 17.9)  \\ \bottomrule
 \end{tabular}
\end{sc}
\end{small}
\end{center}
\vskip -0.1in
\end{table}

 \begin{figure}[H]
\vskip 0.2in
\begin{center}
\centerline{\includegraphics[width=\textwidth]{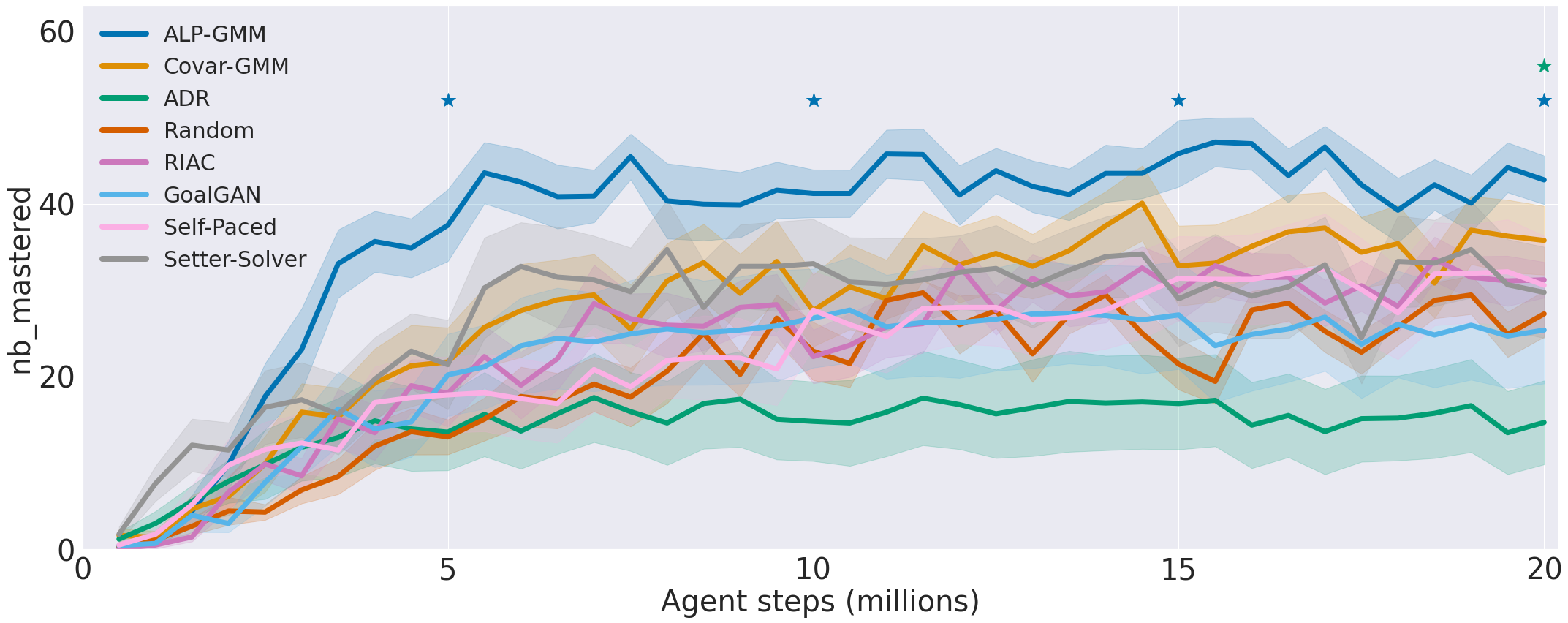}}
\caption{Average percentage of mastered tasks over $16$ seeds using the \textbf{bipedal walker} along with the standard error of the mean. We calculate every $5$ millions steps which method obtained statistically different ($p<0.05$) results from Random and indicate it with a star.}
\label{fig:parkour_bipedal}
\end{center}
\vskip -0.2in
\end{figure}
 
\begin{figure}[H]
\vskip 0.2in
\begin{center}
\centerline{\includegraphics[width=\textwidth]{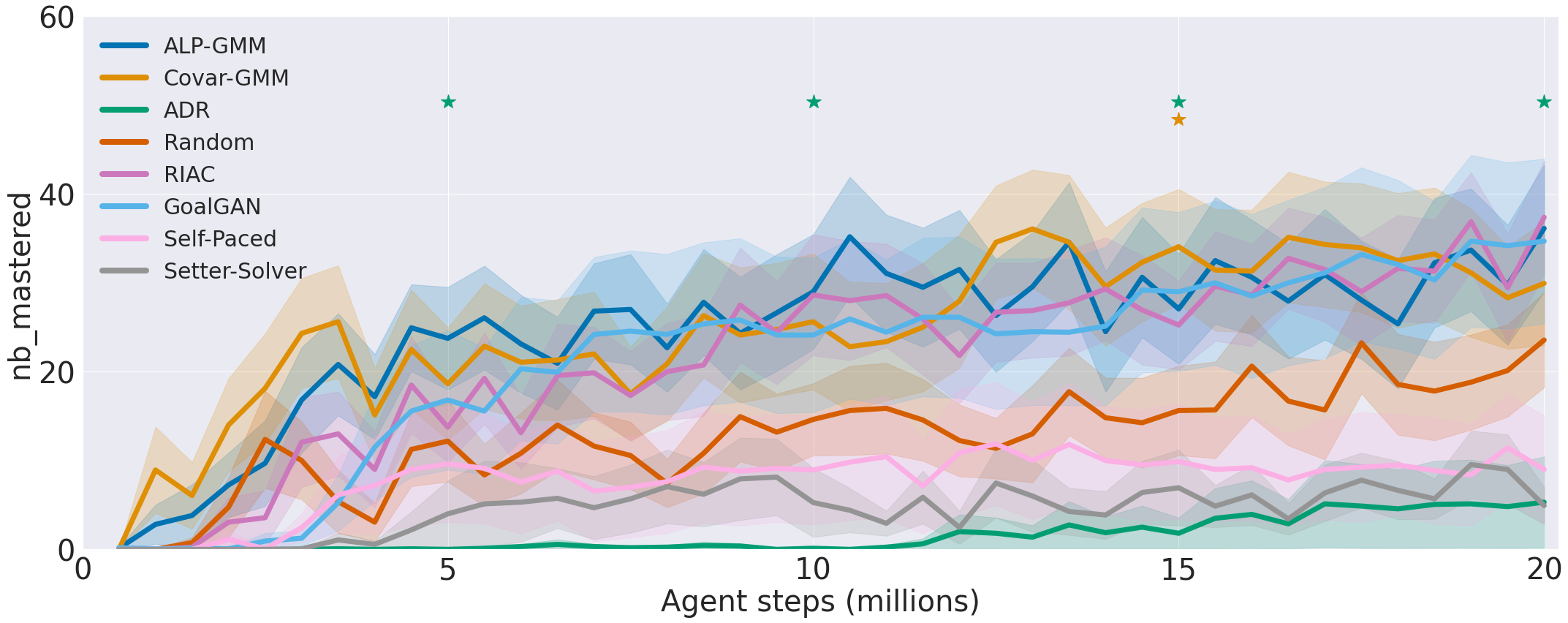}}
\caption{Average percentage of mastered tasks over $16$ seeds using the \textbf{fish} along with the standard error of the mean. We calculate every $5$ millions steps which method obtained statistically different ($p<0.05$) results from Random and indicate it with a star.}
\label{fig:parkour_fish}
\end{center}
\vskip -0.2in
\end{figure}
 
\begin{figure*}[ht]
\vskip 0.2in
\begin{center}
\centerline{\includegraphics[width=\textwidth]{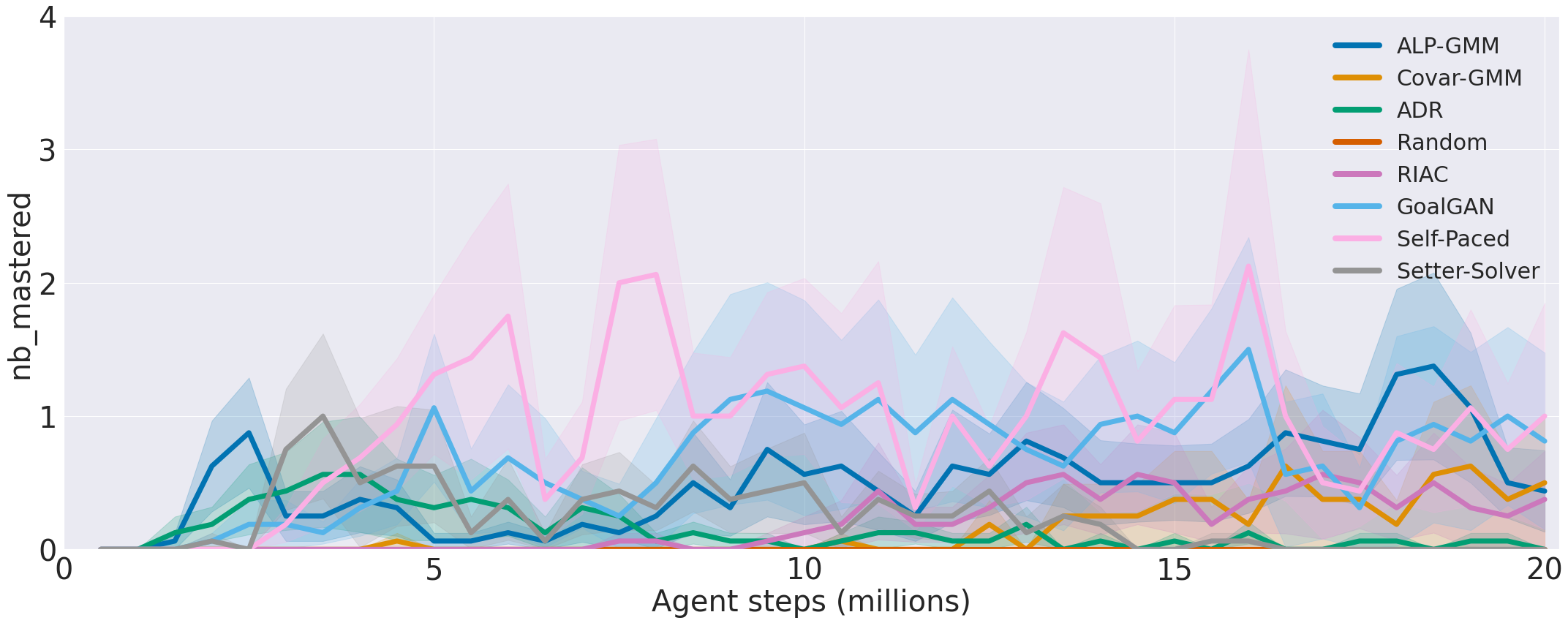}}
\caption{Average percentage of mastered tasks over $16$ seeds using the \textbf{chimpanzee} along with standard error of the mean. We calculate every $5$ millions steps which method obtained statistically different ($p<0.05$) results from Random and indicate it with a star.}
\label{fig:parkour_chimpanzee}
\end{center}
\vskip -0.2in
\end{figure*}

\subsubsection{Case study: Learning climbing locomotion}
\begin{figure}[H]
\vskip 0.2in
\begin{center}
\centerline{\includegraphics[width=\textwidth]{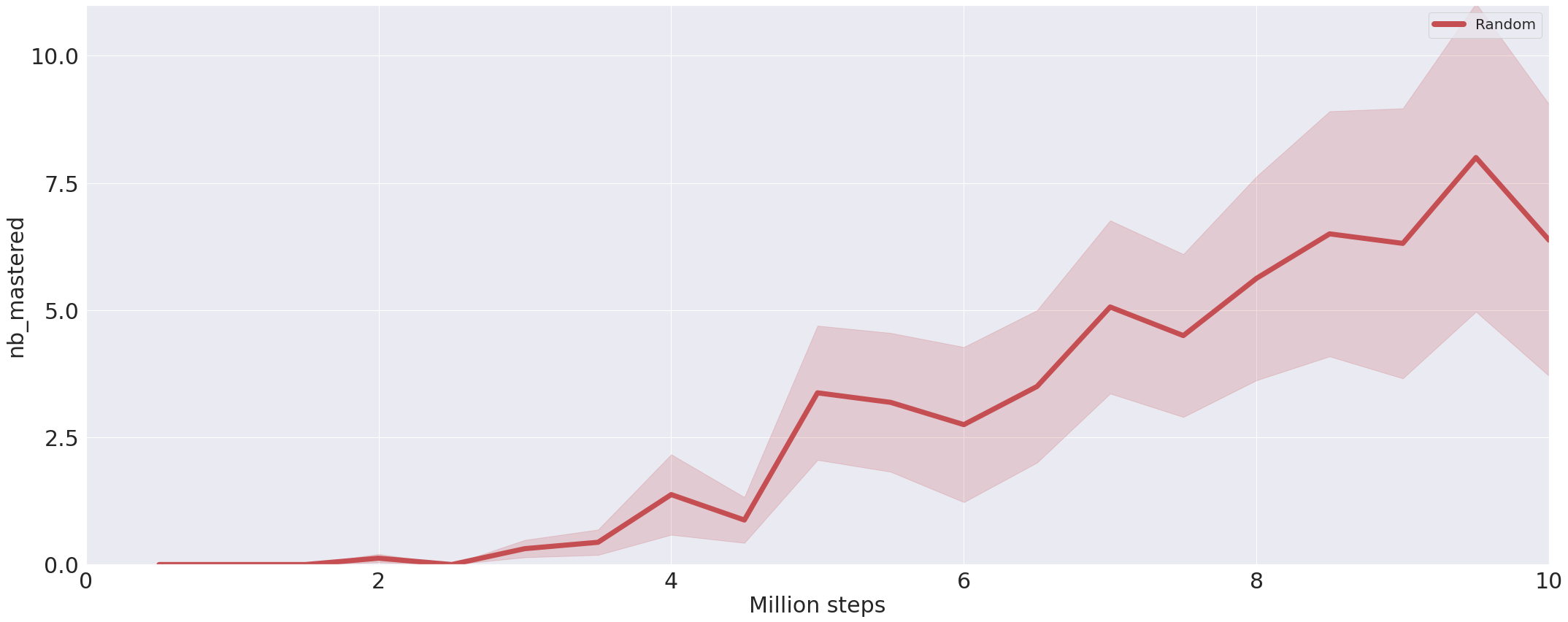}}
\caption{\textbf{Random teacher in the easy CPPN's input space with no water}. Average percentage of mastered tasks over $16$ seeds using our chimpanzee embodiment along with the standard error of the mean.}
\label{fig:parkour_chimpanzee_easy}
\end{center}
\vskip -0.2in
\end{figure}

As shown in figures \ref{fig:parkour_comparison} and \ref{fig:parkour_chimpanzee}, none of the ACL methods implemented in \benchname~ managed to find a curriculum helping the student to learn an efficient climbing policy and master more than $1\%$ of our test set. While learning climbing locomotion can arguably appear as a harder challenge compared to the swimming and walking locomotion, we present in this case study the results of an experiment using our easy CPPN's input space (see appendix \ref{app:expe-details-parkour}), as well as no water (i.e. the maximum level is set to $0.2$, leading to no tasks with water). Using this, we show that simplifying the task space allows our Random teacher to master more than $6\%$ our test set with its best seed reaching $30\%$ at the end of learning in only $10$ millions steps. In comparison, our results in the benchmark show a best performance of $1\%$ of mastered tasks (SPDL) with its best seed reaching only $14\%$ by the end of learning.
As this simpler task space contains more feasible tasks, these results show that the poor performance obtained with the chimpanzee embodiment are due to the inability of the implemented ACL algorithms to find feasible subspaces for their student. This also hints possible better performance by future methods in this totally open challenge of \benchname. See figure \ref{fig:parkour_chimpanzee_easy} for the evolution of percentage of mastered tasks by the Random teacher in this simpler experiment.

